\documentclass{JudgeModel}
\usepackage[utf8]{inputenc} 
\usepackage[T1]{fontenc}    
\usepackage{hyperref}       
\usepackage{url}            
\usepackage{booktabs}       
\usepackage{amsfonts}       
\usepackage{nicefrac}       
\usepackage{microtype}      
\usepackage{marvosym}
\usepackage{soul}
\usepackage{mwe}
\usepackage{mathtools}
\usepackage{graphicx}
\usepackage{threeparttable}
\usepackage{amssymb}
\usepackage{adjustbox}
\usepackage{colortbl}
\usepackage{float}
\usepackage{threeparttable} 
\usepackage{array}
\newcolumntype{P}[1]{>{\centering\arraybackslash}p{#1}}
\usepackage{multirow}
\usepackage{makecell}
\usepackage{bbm}
\usepackage{collcell,xfp}
\usepackage{pgf}
\usepackage{tikz}
\usepackage[most]{tcolorbox}
\usepackage{csquotes}
\usepackage[noorphans,vskip=1em,leftmargin=1em]{quoting}
\usepackage{pifont} 
\usepackage{enumitem} 
\usepackage{tcolorbox} 
\usepackage{xcolor}
\usepackage{forest}
\usepackage{wrapfig}
\usepackage{caption}
\usepackage{subcaption}
\usepackage{longtable}
\usepackage{algpseudocode}
\usepackage{amsmath}
\usepackage[capitalize]{cleveref}
\usepackage[ruled,vlined,linesnumbered]{algorithm2e} 
\usepackage[percent]{overpic}
\usepackage{xspace}
\usepackage[normalem]{ulem}  
\usepackage{tocloft}  
\usepackage{titlesec}
\usepackage[authoryear, round]{natbib}  
\usepackage{tabularx}
\usepackage{listings}
\usepackage{amsthm}
\usepackage{multicol}
\usepackage[table]{xcolor}
\definecolor{heatblue}{RGB}{91,155,213}

\lstdefinestyle{promptstyle}{
    basicstyle=\rmfamily\footnotesize,
    breaklines=true,
    breakatwhitespace=false,
    columns=fullflexible,
    keepspaces=true,
    frame=single,
    backgroundcolor=\color{gray!5},
    rulecolor=\color{gray!40},
    xleftmargin=1em,
    xrightmargin=1em
}
\newcolumntype{C}[1]{>{\centering\arraybackslash}m{#1}}
\newcolumntype{Y}{>{\raggedright\arraybackslash}X}
\newcolumntype{L}[1]{>{\raggedright\arraybackslash}m{#1}}

\usepackage{fontawesome}

\newcommand{\worldwideweb}{\raisebox{-1.5pt}{\includegraphics[height=1.05em]{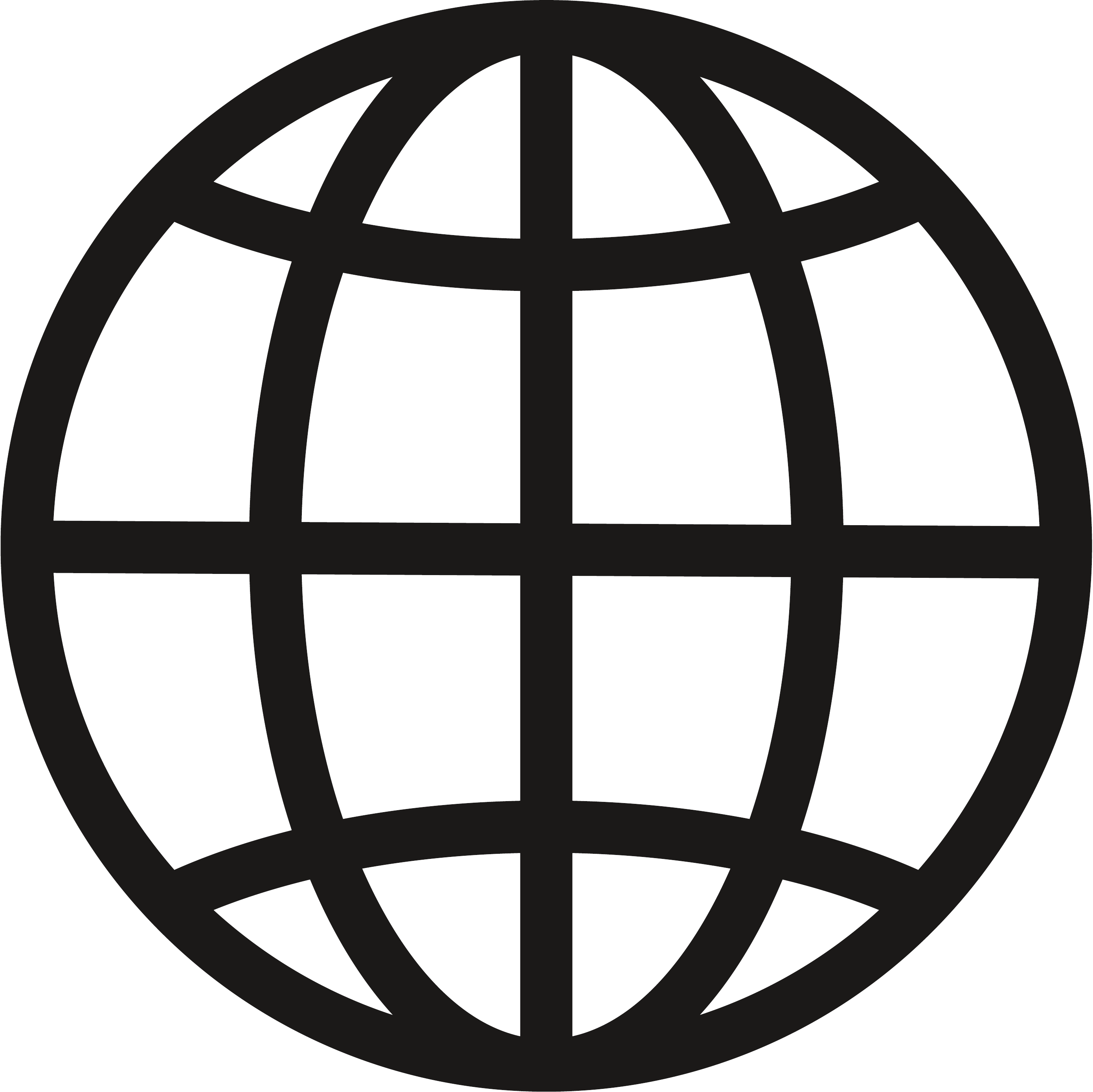}}\xspace}

\definecolor{boxbg}{RGB}{252,248,240}
\definecolor{boxline}{RGB}{188,146,74}

\definecolor{bluelink}{RGB}{0,75,150}
\hypersetup{
    colorlinks=true,%
    citecolor=bluelink,%
    filecolor=bluelink,%
    linkcolor=bluelink,%
    urlcolor=bluelink
}

\newtcolorbox{leaderbox}{
    colback=boxbg,
    colframe=boxline,
    boxrule=0.8pt,
    arc=3mm,
    left=6pt,
    right=6pt,
    top=6pt,
    bottom=6pt,
    enhanced,
}

\newcommand{\runningtitletext}{}
\usepackage{etoolbox}
\newcommand{\currentnavsection}{sec:intro}
\newcommand{\setnavsection}[1]{%
  \gdef\currentnavsection{#1}%
}

\newcommand{\navseclink}[3]{%
  \begingroup
  \edef\currentsec{\currentnavsection}%
  \def\targetsec{#1}%
  \ifx\currentsec\targetsec
    \hyperref[#1]{\textcolor{black}{\textbf{Sec~#2:~#3}}}%
  \else
    \hyperref[#1]{\textcolor{black}{Sec~#2:~#3}}%
  \fi
  \endgroup
}

\newcommand{\navplainlink}[2]{%
  \begingroup
  \edef\currentsec{\currentnavsection}%
  \def\targetsec{#1}%
  \ifx\currentsec\targetsec
    \hyperref[#1]{\textcolor{black}{\textbf{#2}}}%
  \else
    \hyperref[#1]{\textcolor{black}{#2}}%
  \fi
  \endgroup
}

\newcommand{\sectionheadertext}{%
  \small
  \navseclink{sec:intro}{1}{Intro} |
  \navseclink{sec:pipeline}{2}{Method} |
  \navseclink{sec:evaluation}{3}{Eval} |
  \navseclink{sec:findings}{4}{Findings} |
  \navseclink{sec:conclusion}{5}{Conclusion} |
  \navseclink{sec:contributors}{6}{Authors} |
  \navplainlink{sec:references}{References} \\
  \navplainlink{app:related_work}{\textcolor{gray}{Appendix A: Related Work}} |
  \navplainlink{app:metric}{\textcolor{gray}{B: Metrics}} |
  \navplainlink{app:benchmark}{\textcolor{gray}{C: RoboPulse++}} |
  \navplainlink{app:computational_cost}{\textcolor{gray}{D: Efficiency}} |
  \navplainlink{app:robodojo}{\textcolor{gray}{E: Consistency}} |
  \navplainlink{app:more}{\textcolor{gray}{F: Visualization}}
}

\fancypagestyle{mainstyle}{
    \fancyhf{}
    \fancyhead[C]{%
        \vbox{%
            \centering
            {\normalfont\bfseries\fontsize{9.5}{11}\selectfont \runningtitletext\par}%
            \vspace{1pt}
            {\normalfont\fontsize{9.5}{11}\selectfont \sectionheadertext\par}%
            \vspace{3pt}
            {\color{black}\hrule height 0.5pt}%
        }%
    }

    \fancyfoot[C]{\footerfont \thepage}

}

\definecolor{navyblue}{HTML}{0071BC}
\iftrue   
\newcommand{\displaytodo}[1]{#1}
\else
\newcommand{\displaytodo}[1]{}
\fi

\definecolor{blindcolor}{HTML}{AB2AC6}    
\definecolor{chancecolor}{HTML}{F59E0B}   
\definecolor{singlecolor}{HTML}{06B6D4}   
\definecolor{multiplecolor}{HTML}{2563EB} 
\definecolor{captioncolor}{HTML}{22C55E}  

 \newcommand{\culine}[2]{%
    \def\temp@uline{\bgroup\markoverwith
        {\textcolor{#1}{\rule[-0.5ex]{2pt}{1pt}}}\ULon}%
    \temp@uline{#2}%
}
 \newcommand{\cthickuline}[3][0.8pt]{%
    \def\temp@uline{\bgroup\markoverwith
        {\textcolor{#2}{\rule[-0.5ex]{2pt}{#1}}}\ULon}%
    \temp@uline{#3}%
}

\title{
\smash{\raisebox{-0.2\height}{\includegraphics[height=1.3cm]{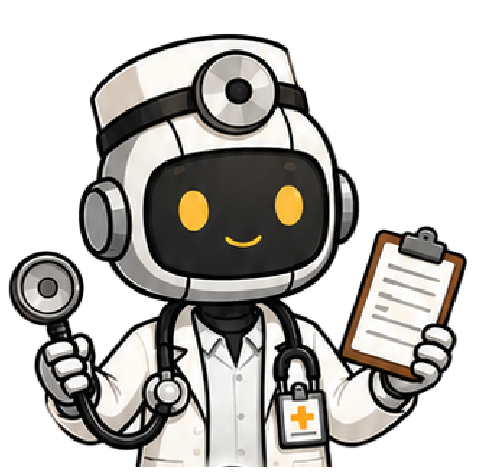}}}%
\centering  \ \ PRM-as-a-Judge 1.5 \\[0.4em]
\hspace{1em}A Toolkit for Robot Process Assessment
}

\renewcommand\Affilfont{\normalfont\fontsize{11}{15}\selectfont\centering}
\author{%
  PRM-as-a-Judge Team
}

\newcommand{\linktable}{
\begin{center}
    \renewcommand{\arraystretch}{1}
    \begin{tabular}{@{}r@{\hspace{0.5em}}l@{\hspace{1.2em}}l@{}}
        \worldwideweb{} & \textbf{Website} & \url{https://prm-as-a-judge.github.io}\\
    \end{tabular}
\end{center}
}

\begin{abstract}
Fine-grained robotic evaluation matters for understanding embodied models, going beyond binary success rates and rule-based process scores.
We present \textbf{PRM-as-a-Judge 1.5}, a toolkit for robot process assessment that turns rollout videos into dense progress curves and derives multiple fine metrics.
PRM-as-a-Judge 1.5 introduces three metrics, building on version 1.0, that characterize failure-side progress, post-drawdown recovery, and success-side execution quality, helping users understand embodied model capability.
Based on the rollout videos from benchmarks, we perform a comprehensive assessment of the embodied models, providing some fine-grained metric results and key findings.
We further introduce RoboPulse++ to evaluate the reliability of process reward models (PRM), providing evaluators with a more accurate testing platform.
Moreover, we release a user-friendly assessment suite, including the benchmark, metric implementation, and visualization tools, to support reproducible manipulation process evaluation.
We call on the community to rethink how robots are evaluated and establish transparent, procedural, and reproducible assessment as a foundation for the next generation of embodied intelligence.
\end{abstract}

\newcommand{\abstractboxwidth}{0.84\linewidth}
\makeatletter
\renewcommand{\abscontent}{%
    \begin{center}
        {\color{black}\fontsize{15pt}{14pt}\selectfont\textbf{Abstract}\par}%
        \vspace{1ex}%
        \parbox{\abstractboxwidth}{\absfont \theabstract}%
        \@ifundefined{@keywords}{}{%
            \vskip1em
            \parbox{\abstractboxwidth}{\keywordsfont Keywords: \@keywords}%
        }%
    \end{center}%
}
\makeatother

\begin{document}
\pagestyle{mainstyle}
\let\storedabscontent\abscontent

\maketitle

\vspace{2em}
\begin{figure}[!htbp]
    \centering
    \includegraphics[width=\linewidth]{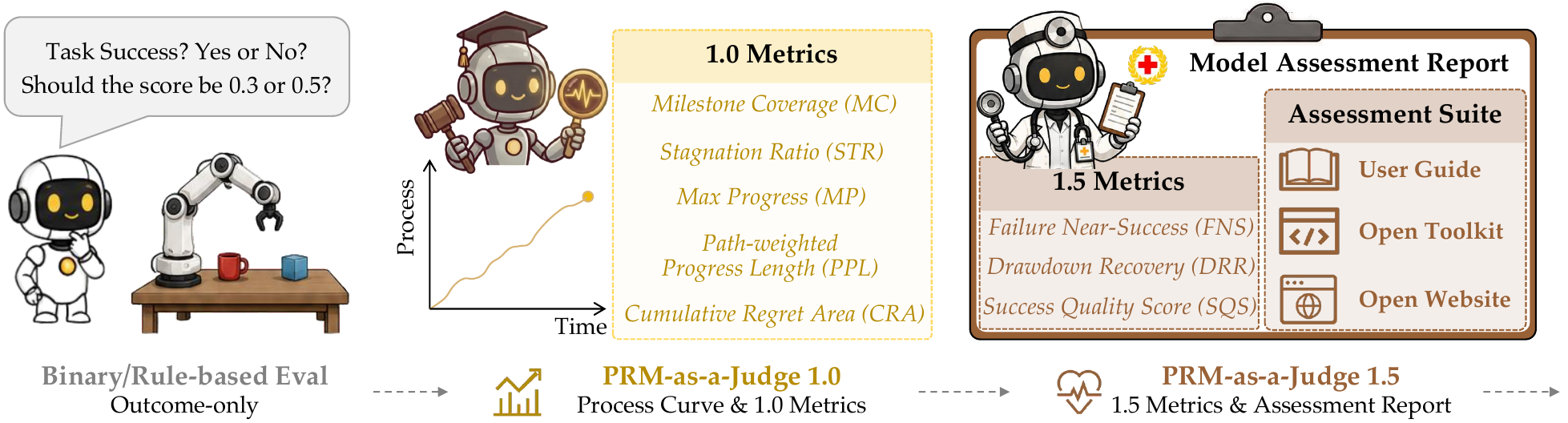}
    \caption{\textbf{The evolution of robotic metrics.} Evaluation progresses from coarse binary outcomes to holistic process understanding. PRM-as-a-Judge 1.0 introduces multi-dimensional process assessment using OPD metrics. Building upon this foundation, PRM-as-a-Judge 1.5 not only evaluates execution quality but also produces a \textit{Model Assessment Report} for embodied models, including VLAs and WAMs.} 
    \label{fig:teaser}
\end{figure}

\section{Introduction}
\label{sec:intro}
\setnavsection{sec:intro}

As embodied models advance from short, isolated manipulation toward long, multi-task environments. Execution becomes increasingly complex, and failure patterns become more diverse. Therefore, evaluation can no longer rely solely on final task success. It must reveal how models evolve throughout progress, where they hesitate, when they regress, and whether they can recover from unstable states. 

Most robotic manipulation benchmarks are dominated by \textit{\textbf{binary success rates}} \citep{mees2022calvin, libero, li2024simplerenv, robotwin20, LIBERO-Plus, VLABench}. Recent works have attempted to expand the metrics to \textit{\textbf{rule-based score manually}} \citep{robochallenge, robodojo}.
These metrics help report whether a task is completed, but they reduce complex trajectories to coarse overall success rates or scores, thus ignoring the fine-grained execution quality.
Intuitively:

\vspace{-1ex}
\begin{enumerate}[label={}, leftmargin=10pt, itemindent=0pt, itemsep=0.5em, topsep=0.5em]
    \item[] \textit{1) Failed rollouts may exhibit very different levels of capability.} \ Even in failed rollouts, some rollouts may fail entirely in the early stages, while others may complete most of their tasks before failing.
    \item[] \textit{2) Successful rollouts vary significantly in quality.} \ Even in successful rollouts, some rollouts achieve the goal smoothly, while others rely on inefficient corrections, hesitation, or recovery behaviors.
\end{enumerate}

The \textit{\textbf{binary success rates}} or \textit{\textbf{rule-based score manually}} are insufficient for assessing embodied model capability, understanding failure modes, or guiding further embodied model development.

We present \textbf{PRM-as-a-Judge 1.5}, which upgrades the progress-curve assessment in PRM-as-a-Judge 1.0 \citep{PRM-as-a-Judge} to a doctor-style model assessment.
Figure~\ref{fig:teaser} summarizes the evolution of robotic metrics and the central idea of PRM-as-a-Judge 1.5. 
This system converts rollout videos into progress curves through process reward models (PRM), computes process metrics, and generates assessment reports for fine-grained analysis. 
PRM-as-a-Judge 1.5 continues the OPD (Outcome–Process–Diagnosis) metric system and introduces three conditioned metrics that separately characterize near-success failures, post-regression recovery, and success-conditioned execution quality.
Beyond metrics, PRM-as-a-Judge 1.5 also releases an assessment suite, including the toolkit codebase and user manual for ease of use.

Based on the available rollout videos from mainstream benchmarks~\citep{robodojo}, we conducted a comprehensive evaluation of these embodied models. This includes performance in terms of reachability, failure-side progress, recovery behavior, success-side quality, and failure fingerprints, and we present some key findings. Furthermore, to evaluate the PRM used in our pipeline, we introduce \textbf{RoboPulse++}, a benchmark of intervals sampled from robot rollouts. 
It tests whether PRMs can distinguish increasing and decreasing task progress throughout robot manipulation execution.

We release PRM-as-a-Judge 1.5 as an open-source suite, which lowers the barrier to reproducible process evaluation and supports more transparent, fair, and process-aware evaluation of embodied models. 

In summary, PRM-as-a-Judge 1.5 provides the community with:

\begin{itemize}[leftmargin=2em, itemsep=0.5em, topsep=0.5em]
    \item \textit{An expanded metric system.} \ It covers reachability, efficiency, stagnation, failure-side progress, recovery, and success-side quality, enabling users to interpret the embodied model capability.
    \item \textit{A comprehensive assessment of mainstream embodied models.} \ PRM-as-a-Judge 1.5 assesses large-scale rollout benchmarks, giving a series of insights beyond official success-rate rankings.
    \item \textit{A RoboPulse++ to guarantee PRM.} \ To demonstrate that PRM provides a more accurate process curve, we introduce RoboPulse++, an interval-level evaluator, to evaluate PRM, providing the community with a more robust testing platform to compare process evaluators.
    \item \textit{An open-source assessment suite.} \  We provide the benchmark, metric implementation, visualization tools, and guide, lowering the barrier to reproducible robot process assessment.
\end{itemize}

\section{PRM-as-a-Judge 1.5}
\label{sec:pipeline}
\setnavsection{sec:pipeline}

PRM-as-a-Judge 1.5 provides an end-to-end assessment framework for embodied models. 
Given the rollout video as input, it produces process metrics and the assessment report for this model.

\subsection{Pipeline}
\label{subsec:pipeline}
\begin{figure}[!htbp]
    \centering
    \includegraphics[width=\linewidth]{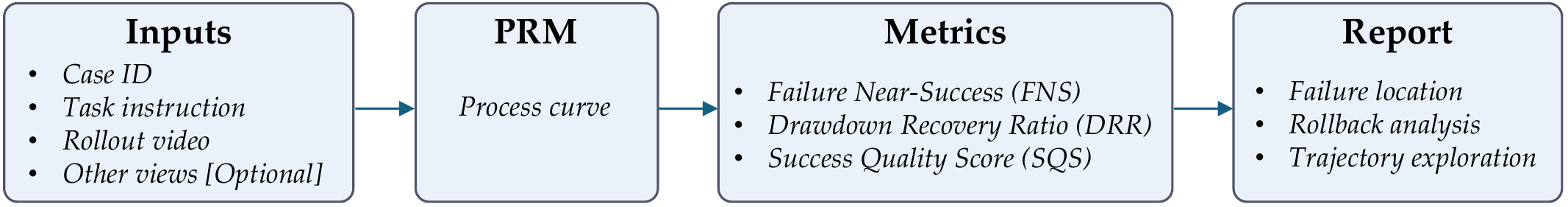}
    \caption{\textbf{Overall pipeline of PRM-as-a-Judge 1.5.} After receiving the rollout video and related inputs, PRM-as-a-Judge 1.5 first uses the PRM to obtain a progress curve. Additionally, we update the metrics of version 1.0. Based on this progress curve and the designed metrics, users can obtain a comprehensive report. This report includes fine-grained evaluation and assessment results of the embodied models.}
    \label{fig:pipeline}
\end{figure}

Figure~\ref{fig:pipeline} shows the overall pipeline of PRM-as-a-Judge 1.5. It receives rollout videos, estimates progress curves with PRM, computes process metrics, and generates reports for the embodied model assessment.

\paragraph{Inputs.}
PRM-as-a-Judge 1.5 organizes the rollout into a record containing a unique \textit{Case ID}, \textit{Task instruction}, and \textit{Rollout video}. Users can also optionally provide \textit{Other views} as additional observations.

\paragraph{PRM.}\label{subsec:prm} 
PRM estimates task completion at each frame or timestamp to produce the \textit{Process curve} for the rollout.
PRMs typically use paired or sequential inputs, such as Robo-Dopamine~\citep{robodopamine} and RoboReward \citep{RoboReward}, which compare two frames, and RoboMeter~\citep{liang2026robometer} and RynnValue \citep{RynnValue}, which jointly process multiple frames.

\paragraph{Metrics.}
This step translates the \textit{Process curve} into detailed metrics for the embodied model assessment. 
To describe the different aspects of execution, 
PRM-as-a-Judge 1.5 improves upon the OPD metrics of PRM-as-a-Judge 1.0 \citep{PRM-as-a-Judge}, adding three conditioned metrics such as \textit{Failure Near-Success (FNS)}, \textit{Drawdown Recovery Ratio (DRR)}, and \textit{Success Quality Score (SQS)}. These provide an assessment of effectiveness, reasons for failure, and degree of recovery. The details of the added metrics in PRM-as-a-Judge 1.5 can be seen in \textbf{Section \ref{subsec:method}}.

\paragraph{Report.}
The final assessment report will include features such as \textit{Failure location}, \textit{Rollback analysis}, and \textit{Trajectory exploration}, synchronizing \textit{Rollout video} with the corresponding \textit{Progress curve}. Users can obtain results, such as early failure, near success, stagnation, recovery, and so on. For more reports, please see \textbf{Section \ref{sec:evaluation}} and \textbf{Appendix \ref{subsec:evaluation_visualization}}. Additionally, users can view executable examples in the user guide.
\subsection{Metrics}\label{subsec:method}

\paragraph{OPD system.} The core idea is to construct a comprehensive metric system to cover the embodied model's progress, reasons for failure, and execution quality.
We introduce the OPD (Outcome–Process-Diagnosis) system: \textit{Outcome} describes stage-wise reachability, \textit{Process} captures progress efficiency, and \textit{Diagnosis} quantifies regression and stagnation patterns that expose failure and recovery behavior. 

PRM-as-a-Judge 1.5 further introduces three conditioned diagnosis-level metrics, making OPD metrics more actionable for embodied model assessments.
Overall, these metrics turn a progress curve into an assessment. Table~\ref{tab:metrics} provides the metrics in PRM-as-a-Judge 1.5. They can show how the embodied model progresses, where it fails, whether it can recover, and its reliability in completing the task.

\begin{table}[!ht]
\centering
\footnotesize
\setlength{\tabcolsep}{2.5pt}
\renewcommand{\arraystretch}{1.4}
\caption{\textbf{Overview of the OPD Metric Suite.} Pref. is Preference direction ($\uparrow$: higher is better, and vice versa). $\dag$ represents new metrics added in PRM-as-a-Judge 1.5.
}
\label{tab:metrics}
\begin{tabularx}{\linewidth}{
>{\raggedright\arraybackslash}m{4.7cm}
>{\centering\arraybackslash}m{0.6cm}
>{\raggedright\arraybackslash}m{6cm}
>{\raggedright\arraybackslash}m{4cm}
}
\toprule
\textbf{Metric} & \textbf{Pref.} & \textbf{Definition} & \textbf{Interpretation} \\
\midrule[0.5pt]
\rowcolor[rgb]{ .949,  .949,  .949}\multicolumn{4}{l}{\textit{\textbf{Outcome–level Metrics}}} \\
Milestone Coverage (MC@\(q\)) & $\uparrow$ & The proportion of rollouts whose maximum progress reaches milestone \(q\). & 
Reveal where embodied models stop progressing. \\
Max Progress (MP) & $\uparrow$ & The maximum progress value attained during the whole rollout. & 
The furthest progress reached. \\
\midrule[0.5pt]
\rowcolor[rgb]{ .949,  .949,  .949}\multicolumn{4}{l}{\textit{\textbf{Process–level Metrics}}} \\
Path-weighted Progress Length (PPL) & $\uparrow$ & Squared maximum progress divided by the total variation of the progress curve. & Measure how efficiently a rollout reaches its best progress. \\
\midrule[0.5pt]
\rowcolor[rgb]{.949, .949, .949}\multicolumn{4}{l}{\textit{\textbf{Diagnosis- level Metrics}}} \\
Cumulative Regret Area (CRA) & $\downarrow$ & The average gap between the process at each timestep and the highest process reached before. & The severity and duration of regression from the best-so-far progress level. \\
Stagnation Ratio (STR)& $\downarrow$ & The proportion of consecutive timesteps whose progress change falls below a noise threshold. & How often execution stalls or hesitates. \\
Failure Near-Success (FNS) {$^ \dag$} & $\uparrow$ & A composite of MP, MC@50, and MC@75 for failed rollouts. &  How close a failed rollout came to completion. \\
Drawdown Recovery Ratio (DRR) {$^ \dag$} & $\uparrow$ & The largest subsequent recovery divided by the maximum drawdown for rollouts that experience a drawdown.
& How much of the largest setback is recovered.\\
Success Quality Score (SQS) {$^ \dag$} & $\uparrow$ & A composite of PPL, CRA, and STR for successful rollouts. & The efficiency and stability of successful execution. \\
\bottomrule
\end{tabularx}
\end{table}
\section{Evaluation}\label{sec:evaluation}
\setnavsection{sec:evaluation}

In this section, we validate PRM-as-a-Judge 1.5 on embodied model assessment across the manipulation tasks. 
Beyond binary success rates, PRM-as-a-Judge 1.5 reveals differences in the evaluation: \textit{\textbf{how far the model progresses, how efficiently it executes, and whether it regresses, stagnates, or recovers}}.
We use RoboDopamine (Forward) as the default judge model, and evaluation details about judge models can be found in the \textbf{Appendix~\ref{app:benchmark}}.
Note: the analysis and findings in this report are derived from the rollout videos released by the corresponding benchmarks. We do not fine-tune or otherwise modify the evaluated embodied models.
To further avoid potential benchmark-specific optimizations for RoboDojo, we restrict our evaluation to models in the leaderboard that is frozen on \textit{3 Jul. 2026}.

\subsection{\textbf{Settings}}
\paragraph{Benchmark.} We use RoboDojo~\citep{robodojo}, a benchmark covering both real-world (\textbf{\textit{RoboDojo-RealWorld}}) and simulation (\textbf{\textit{RoboDojo-Sim}}) tasks.
RoboDojo-RealWorld evaluates challenging deployment. RoboDojo-Sim evaluates generalization, memory, long-horizon results, and instruction following.

\paragraph{Baselines.} In the \textbf{\textit{RoboDojo-RealWorld}}, we analyze these models (GR00T-N1.7~\citep{GR00TN1.7}, GalaxeaVLA~\citep{galaxea2025}, InternVLA-A1~\citep{internvla}, $\pi_0$~\citep{pi0}, $\pi_{0.5}$~\citep{pi0.5}, Spirit v1.5~\citep{spirit}, StarVLA-$\alpha$~\citep{StarVLA}, X-VLA~\citep{xvla}, and Xiaomi-Robotics-0~\citep{xiaomi0}). 
In the \textbf{\textit{RoboDojo-Sim}}, we analyze these embodied models (Hy-Embodied-0.5-VLA~\citep{hy}, Xiaomi-Robotics-1~\citep{xiaomi1}, Spatial Forcing (w/ $\pi_{0.5}$)~\citep{sf}, X-WAM~\citep{xwam}, GigaWorld-Policy-0~\citep{GigaWorld}, AHA-WAM~\citep{aha}, Fast-WAM~\citep{fastwam}, and LDA-1B~\citep{lda}).

\subsection{Leaderboard}
\vspace{-5pt}
Table~\ref{tab:comparison_dojo_real} and Table~\ref{tab:comparison_dojo_sim} report the evaluation performance on RoboDojo-RealWorld and RoboDojo-Sim.

\begin{table}[!htbp]
\centering
\small
\setlength{\tabcolsep}{1.2mm}
\begin{threeparttable}
\caption{\textbf{Comparisons on RoboDojo-RealWorld.} $\uparrow$ represents a higher value, the better, the darker the color, and vice versa. \textbf{Bold} represents the best results, and \underline{underline} represents the suboptimal results.} 
\label{tab:comparison_dojo_real}
\begin{tabular}{c|ccccccccccc}
\toprule[1pt]
\textbf{\textit{Embodied Model}} & MC@25 $\uparrow$\tnote{1} & MC@50 $\uparrow$\tnote{1} & MC@75 $\uparrow$\tnote{1} & MP $\uparrow$\tnote{2} & SR $\uparrow$\tnote{3} & PPL $\uparrow$\tnote{4} & DRR $\uparrow$\tnote{5} & FNS $\uparrow$\tnote{6} & SQS $\uparrow$\tnote{7} & CRA $\downarrow$\tnote{8} & STR $\downarrow$\tnote{9} \\
\midrule[0.5pt]
$\pi_{0.5}$ & \cellcolor{heatblue!50}\textbf{85.29} & \cellcolor{heatblue!50}\textbf{60.59} & \cellcolor{heatblue!50}\textbf{38.82} & \cellcolor{heatblue!50}\textbf{59.90} & \cellcolor{heatblue!50}\textbf{17.06} & \cellcolor{heatblue!50}\textbf{28.93} & \cellcolor{heatblue!50}\textbf{100.00} & \cellcolor{heatblue!50}\textbf{45.39} & \cellcolor{heatblue!45}82.08 & \cellcolor{heatblue!50}\textbf{5.22} & \cellcolor{heatblue!50}\textbf{21.81} \\
InternVLA-A1  & \cellcolor{heatblue!30}61.76 & \cellcolor{heatblue!25}\underline{33.53} & \cellcolor{heatblue!25}\underline{18.82} & \cellcolor{heatblue!20}30.42 & \cellcolor{heatblue!20}\underline{4.71} & \cellcolor{heatblue!20}14.18 & \cellcolor{heatblue!25}66.81 & \cellcolor{heatblue!15}14.83 & \cellcolor{heatblue!50}86.62 & \cellcolor{heatblue!35}\underline{9.24} & \cellcolor{heatblue!25}38.15 \\
Xiaomi-Robotics-0  & \cellcolor{heatblue!25}54.71 & \cellcolor{heatblue!20}29.41 & \cellcolor{heatblue!15}11.76 & \cellcolor{heatblue!20}27.12 & \cellcolor{heatblue!20}3.53 & \cellcolor{heatblue!15}10.50 & \cellcolor{heatblue!20}62.70 & \cellcolor{heatblue!15}13.14 & \cellcolor{heatblue!45}84.47 & \cellcolor{heatblue!20}13.01 & \cellcolor{heatblue!30}35.45 \\
GalaxeaVLA (G0) & \cellcolor{heatblue!30}60.00 & \cellcolor{heatblue!20}27.65 & \cellcolor{heatblue!15}10.00 & \cellcolor{heatblue!20}29.41 & \cellcolor{heatblue!15}2.35 & \cellcolor{heatblue!15}11.17 & \cellcolor{heatblue!25}66.69 & \cellcolor{heatblue!15}14.36 & \cellcolor{heatblue!50}88.21 & \cellcolor{heatblue!25}11.87 & \cellcolor{heatblue!30}35.48 \\
X-VLA & \cellcolor{heatblue!35}\underline{69.41} & \cellcolor{heatblue!25}32.35 & \cellcolor{heatblue!25}17.06 & \cellcolor{heatblue!30}\underline{37.21} & \cellcolor{heatblue!15}2.35 & \cellcolor{heatblue!15}11.38 & \cellcolor{heatblue!25}66.31 & \cellcolor{heatblue!20}\underline{18.24} & \cellcolor{heatblue!45}83.45 & \cellcolor{heatblue!10}15.76 & \cellcolor{heatblue!50}\underline{22.39} \\
$\pi_{0}$ & \cellcolor{heatblue!25}54.12 & \cellcolor{heatblue!15}21.18 & \cellcolor{heatblue!10}7.65 & \cellcolor{heatblue!20}26.99 & \cellcolor{heatblue!15}1.76 & \cellcolor{heatblue!15}10.69 & \cellcolor{heatblue!25}\underline{71.27} & \cellcolor{heatblue!15}12.96 & \cellcolor{heatblue!50}\underline{88.57} & \cellcolor{heatblue!30}10.15 & \cellcolor{heatblue!25}38.80 \\
StarVLA-$\alpha$ & \cellcolor{heatblue!10}37.65 & \cellcolor{heatblue!10}15.29 & \cellcolor{heatblue!10}7.06 & \cellcolor{heatblue!10}17.94 & \cellcolor{heatblue!15}1.76 & \cellcolor{heatblue!10}8.31 & \cellcolor{heatblue!15}57.04 & \cellcolor{heatblue!10}8.81 & \cellcolor{heatblue!50}86.06 & \cellcolor{heatblue!30}9.93 & \cellcolor{heatblue!10}47.72 \\
GR00T-N1.7 & \cellcolor{heatblue!30}63.75 & \cellcolor{heatblue!20}29.38 & \cellcolor{heatblue!20}13.13 & \cellcolor{heatblue!25}31.64 & \cellcolor{heatblue!10}0.63 & \cellcolor{heatblue!25}\underline{15.60} & \cellcolor{heatblue!15}56.34 & \cellcolor{heatblue!20}15.75 & \cellcolor{heatblue!50}\textbf{91.65} & \cellcolor{heatblue!30}10.15 & \cellcolor{heatblue!25}39.41 \\
Spirit v1.5 & \cellcolor{heatblue!35}66.67 & \cellcolor{heatblue!20}29.17 & \cellcolor{heatblue!10}7.50 & \cellcolor{heatblue!20}29.70 & \cellcolor{heatblue!10}0.00 & \cellcolor{heatblue!15}11.72 & \cellcolor{heatblue!10}50.81 & \cellcolor{heatblue!15}14.85 & \cellcolor{heatblue!10}0.00 & \cellcolor{heatblue!10}15.20 & \cellcolor{heatblue!40}26.89 \\
\bottomrule[1pt]
\end{tabular}
\begin{tablenotes}
\footnotesize
\item[1] MC@p: \textit{Milestone Completion at p\%.} \ They are the proportions of trajectories reaching 25\%, 50\%, and 75\% progress.
\item[2] MP: \textit{MaxProcess.} \ It represents the max-level execution process.
\item[3] SR: \textit{Binary Success Rate.} \ The unit is \%.
\item[4] PPL: \textit{Path-weighted Progress Length.} \ It measures how directly the trajectory reaches its best state.
\item[5] DRR: \textit{Drawdown Recovery Ratio.} \ It measures recovery after the largest progress drawdown and is computed only over trajectories that experience a drawdown.
\item[6] FNS: \textit{Failure Near-Success Score.} \ It measures the progress achieved by failed trajectories before termination.
\item[7] SQS: \textit{Success Quality Score.} \ It measures the stability, smoothness, and quality in the success process.
\item[8] CRA: \textit{Cumulative Regret Area.} \ It measures the persistence of state regression during execution.
\item[9] STR: \textit{Stagnation Ratio.} \ It measures the proportion of time steps where the potential change falls below a noise threshold.
\end{tablenotes}
\end{threeparttable}
\end{table}

\begin{table}[!ht]
\centering
\small
\setlength{\tabcolsep}{0.85mm}
\begin{threeparttable}
\caption{\textbf{Comparisons on RoboDojo-Sim.} Specific identification is the same as Table~\ref{tab:comparison_dojo_real}.}
\label{tab:comparison_dojo_sim}
\begin{tabular}{c|ccccccccccc}
\toprule[1pt]
\textbf{\textit{Embodied Model}} & MC@25 $\uparrow$\tnote{1} & MC@50 $\uparrow$\tnote{1} & MC@75 $\uparrow$\tnote{1} & MP $\uparrow$\tnote{2} & SR $\uparrow$\tnote{3} & PPL $\uparrow$\tnote{4} & DRR $\uparrow$\tnote{5} & FNS $\uparrow$\tnote{6} & SQS $\uparrow$\tnote{7} & CRA $\downarrow$\tnote{8} & STR $\downarrow$\tnote{9} \\
\midrule[0.5pt]
Hy-Embodied-0.5-VLA  & \cellcolor{heatblue!45}67.01 & \cellcolor{heatblue!50}\textbf{46.53} & \cellcolor{heatblue!50}\underline{28.82} & \cellcolor{heatblue!50}\textbf{47.68} & \cellcolor{heatblue!50}\textbf{11.46} & \cellcolor{heatblue!50}\textbf{18.81} & \cellcolor{heatblue!40}83.06 & \cellcolor{heatblue!45}19.99 & \cellcolor{heatblue!25}77.77 & \cellcolor{heatblue!50}\textbf{9.75} & \cellcolor{heatblue!30}35.80 \\
$\pi_{0.5}$ & \cellcolor{heatblue!50}\underline{73.61} & \cellcolor{heatblue!50}\underline{45.83} & \cellcolor{heatblue!45}27.43 & \cellcolor{heatblue!50}\underline{46.60} & \cellcolor{heatblue!40}\underline{8.68} & \cellcolor{heatblue!45}16.88 & \cellcolor{heatblue!45}\underline{87.46} & \cellcolor{heatblue!50}\textbf{21.03} & \cellcolor{heatblue!25}78.76 & \cellcolor{heatblue!50}10.18 & \cellcolor{heatblue!45}30.36 \\
Spatial Forcing (w/ $\pi_{0.5}$) & \cellcolor{heatblue!45}70.83 & \cellcolor{heatblue!50}44.44 & \cellcolor{heatblue!50}\textbf{29.17} & \cellcolor{heatblue!45}44.88 & \cellcolor{heatblue!40}\underline{8.68} & \cellcolor{heatblue!45}\underline{16.98} & \cellcolor{heatblue!50}\textbf{93.47} & \cellcolor{heatblue!45}\underline{20.15} & \cellcolor{heatblue!15}74.19 & \cellcolor{heatblue!45}11.45 & \cellcolor{heatblue!45}\underline{29.59} \\
X-WAM & \cellcolor{heatblue!50}\textbf{73.96} & \cellcolor{heatblue!45}41.67 & \cellcolor{heatblue!40}23.26 & \cellcolor{heatblue!40}41.19 & \cellcolor{heatblue!35}7.29 & \cellcolor{heatblue!40}16.03 & \cellcolor{heatblue!35}80.51 & \cellcolor{heatblue!45}19.38 & \cellcolor{heatblue!25}77.86 & \cellcolor{heatblue!45}10.89 & \cellcolor{heatblue!45}30.88 \\
GalaxeaVLA (G0) & \cellcolor{heatblue!30}51.05 & \cellcolor{heatblue!35}31.82 & \cellcolor{heatblue!30}17.13 & \cellcolor{heatblue!25}26.14 & \cellcolor{heatblue!30}6.29 & \cellcolor{heatblue!25}10.15 & \cellcolor{heatblue!30}74.13 & \cellcolor{heatblue!20}11.55 & \cellcolor{heatblue!25}77.07 & \cellcolor{heatblue!35}13.62 & \cellcolor{heatblue!25}38.52 \\
X-VLA & \cellcolor{heatblue!40}62.59 & \cellcolor{heatblue!45}41.61 & \cellcolor{heatblue!40}23.08 & \cellcolor{heatblue!40}39.84 & \cellcolor{heatblue!30}6.29 & \cellcolor{heatblue!35}13.82 & \cellcolor{heatblue!20}65.52 & \cellcolor{heatblue!40}17.64 & \cellcolor{heatblue!35}81.74 & \cellcolor{heatblue!45}11.17 & \cellcolor{heatblue!50}\textbf{28.41} \\
AHA-WAM & \cellcolor{heatblue!35}54.86 & \cellcolor{heatblue!30}28.82 & \cellcolor{heatblue!30}17.36 & \cellcolor{heatblue!25}27.14 & \cellcolor{heatblue!20}4.51 & \cellcolor{heatblue!20}8.71 & \cellcolor{heatblue!30}77.16 & \cellcolor{heatblue!25}13.10 & \cellcolor{heatblue!25}77.21 & \cellcolor{heatblue!30}15.36 & \cellcolor{heatblue!40}32.69 \\
StarVLA-$\alpha$ & \cellcolor{heatblue!30}53.13 & \cellcolor{heatblue!25}25.69 & \cellcolor{heatblue!20}12.50 & \cellcolor{heatblue!25}27.11 & \cellcolor{heatblue!20}4.17 & \cellcolor{heatblue!25}10.06 & \cellcolor{heatblue!30}73.31 & \cellcolor{heatblue!25}13.04 & \cellcolor{heatblue!40}\underline{85.52} & \cellcolor{heatblue!45}11.55 & \cellcolor{heatblue!15}43.28 \\
Xiaomi-Robotics-0 & \cellcolor{heatblue!35}55.21 & \cellcolor{heatblue!35}32.99 & \cellcolor{heatblue!30}18.06 & \cellcolor{heatblue!25}27.71 & \cellcolor{heatblue!20}3.82 & \cellcolor{heatblue!25}10.49 & \cellcolor{heatblue!25}68.94 & \cellcolor{heatblue!30}13.55 & \cellcolor{heatblue!10}71.84 & \cellcolor{heatblue!40}12.08 & \cellcolor{heatblue!30}35.91 \\
$\pi_{0}$ & \cellcolor{heatblue!30}51.74 & \cellcolor{heatblue!25}26.39 & \cellcolor{heatblue!25}13.54 & \cellcolor{heatblue!25}26.15 & \cellcolor{heatblue!15}3.47 & \cellcolor{heatblue!20}8.75 & \cellcolor{heatblue!25}69.68 & \cellcolor{heatblue!25}12.44 & \cellcolor{heatblue!20}74.88 & \cellcolor{heatblue!35}14.92 & \cellcolor{heatblue!30}36.45 \\
GigaWorld-Policy-0 & \cellcolor{heatblue!35}57.64 & \cellcolor{heatblue!35}34.38 & \cellcolor{heatblue!35}20.14 & \cellcolor{heatblue!25}27.75 & \cellcolor{heatblue!15}3.13 & \cellcolor{heatblue!30}11.38 & \cellcolor{heatblue!15}60.02 & \cellcolor{heatblue!30}13.79 & \cellcolor{heatblue!25}77.31 & \cellcolor{heatblue!50}\underline{9.83} & \cellcolor{heatblue!20}41.42 \\
LDA-1B & \cellcolor{heatblue!15}35.76 & \cellcolor{heatblue!10}13.89 & \cellcolor{heatblue!10}5.90 & \cellcolor{heatblue!15}17.42 & \cellcolor{heatblue!15}2.78 & \cellcolor{heatblue!15}5.50 & \cellcolor{heatblue!15}59.25 & \cellcolor{heatblue!15}8.44 & \cellcolor{heatblue!50}\textbf{88.80} & \cellcolor{heatblue!10}21.22 & \cellcolor{heatblue!15}43.30 \\
Fast-WAM & \cellcolor{heatblue!20}42.71 & \cellcolor{heatblue!20}19.44 & \cellcolor{heatblue!20}11.46 & \cellcolor{heatblue!15}19.15 & \cellcolor{heatblue!15}2.43 & \cellcolor{heatblue!20}7.97 & \cellcolor{heatblue!25}72.74 & \cellcolor{heatblue!15}9.41 & \cellcolor{heatblue!10}71.36 & \cellcolor{heatblue!40}12.15 & \cellcolor{heatblue!10}44.45 \\
GR00T-N1.7 & \cellcolor{heatblue!30}50.69 & \cellcolor{heatblue!25}26.74 & \cellcolor{heatblue!20}10.76 & \cellcolor{heatblue!25}25.40 & \cellcolor{heatblue!10}2.08 & \cellcolor{heatblue!25}9.23 & \cellcolor{heatblue!25}68.61 & \cellcolor{heatblue!25}12.40 & \cellcolor{heatblue!40}85.13 & \cellcolor{heatblue!35}14.66 & \cellcolor{heatblue!15}42.93 \\
InternVLA-A1 & \cellcolor{heatblue!25}45.14 & \cellcolor{heatblue!20}20.14 & \cellcolor{heatblue!15}9.72 & \cellcolor{heatblue!20}22.30 & \cellcolor{heatblue!10}2.08 & \cellcolor{heatblue!20}7.47 & \cellcolor{heatblue!10}56.74 & \cellcolor{heatblue!20}10.68 & \cellcolor{heatblue!35}83.19 & \cellcolor{heatblue!30}15.74 & \cellcolor{heatblue!25}38.17 \\
Spirit v1.5 & \cellcolor{heatblue!10}28.82 & \cellcolor{heatblue!10}12.15 & \cellcolor{heatblue!10}5.56 & \cellcolor{heatblue!10}14.98 & \cellcolor{heatblue!10}1.74 & \cellcolor{heatblue!10}4.21 & \cellcolor{heatblue!20}67.40 & \cellcolor{heatblue!10}7.34 & \cellcolor{heatblue!15}72.57 & \cellcolor{heatblue!10}21.64 & \cellcolor{heatblue!10}44.00 \\
\bottomrule[1pt]
\end{tabular}
\begin{tablenotes}
\footnotesize
\item[1] MC@p: \textit{Milestone Completion at p\%.} \ They are the proportions of trajectories reaching 25\%, 50\%, and 75\% progress.
\item[2] MP: \textit{MaxProcess.} \ It represents the max-level execution process.
\item[3] SR: \textit{Binary Success Rate.} \ The unit is \%.
\item[4] PPL: \textit{Path-weighted Progress Length.} \ It measures how directly the trajectory reaches its best state.
\item[5] DRR: \textit{Drawdown Recovery Ratio.} \ It measures recovery after the largest progress drawdown and is computed only over trajectories that experience a drawdown.
\item[6] FNS: \textit{Failure Near-Success Score.} \ It measures the progress achieved by failed trajectories before termination.
\item[7] SQS: \textit{Success Quality Score.} \ It measures the stability, smoothness, and quality in the success process.
\item[8] CRA: \textit{Cumulative Regret Area.} \ It measures the persistence of state regression during execution.
\item[9] STR: \textit{Stagnation Ratio.} \ It measures the proportion of time steps where the potential change falls below a noise threshold.
\end{tablenotes}
\end{threeparttable}
\end{table}

As shown in Table \ref{tab:comparison_dojo_real} and Table \ref{tab:comparison_dojo_sim}, the relative ranking of embodied models varies across different metrics. In particular, the ranking of SR values is not entirely consistent with process metrics such as SQS, DRR, or FNS. This demonstrates that relying solely on SR to evaluate a model is insufficient.

\section{Key Findings}
\label{sec:findings}
\setnavsection{sec:findings}

Specifically, we organize our experiments around the following questions:
\begin{enumerate}[label={}, leftmargin=10pt, itemindent=0pt, itemsep=0.3em, topsep=0.3em]
    \item[] \textit{\textbf{$\cdot$ Question 1: Which is stronger, vision-language-action models (VLAs) or world action models (WAMs)?}}
    \item[] \textit{\textbf{$\cdot$ Question 2: Does a larger model always result in better performance?}}
    \item[] \textit{\textbf{$\cdot$ Question 3: Which model is generally the strongest?}}
    \item[] \textit{\textbf{$\cdot$ Question 4: What tasks are existing embodied models relatively better at?}}
    \item[] \textit{\textbf{$\cdot$ Question 5: Does simulation performance correlate with real-world performance?}}
\end{enumerate}

Next, we will analyze each question one by one and present our findings.

\vspace{-10pt}

\subsection{Which is stronger, VLAs or WAMs?}
\begin{leaderbox}
\textcolor{boxline}{\textbf{Finding 1}: VLAs generally outperform WAMs under different metrics.}
\end{leaderbox}

\begin{table}[!htbp]
\centering
\small
\setlength{\tabcolsep}{0.85mm}
\caption{\textbf{Rankings on RoboDojo-Sim.} Specific values are shown in Table \ref{tab:comparison_dojo_sim}.}
\label{tab:comparison_dojo_sim1}
\begin{tabular}{
c|
*{3}{>{\centering\arraybackslash}m{1.1cm}}*{8}{>{\centering\arraybackslash}m{0.75cm}}
|>{\centering\arraybackslash}m{0.8cm}
}
\toprule[1pt]
\textbf{\textit{Embodied Model}} & MC@25& MC@50& MC@75& MP& SR & PPL& DRR& FNS& SQS& CRA & STR  &\textbf{Avg.}\\
\midrule[0.5pt]
$\pi_{0.5}$  & 2	&2&	3	&2	&2	&3	&2	&1	&6&	3&	3 &1\\
Hy-Embodied-0.5-VLA	&4&	1	&2	&1	&1	&1	&3&	3&	8	&1	&6&2\\
Spatial Forcing (w/ $\pi_{0.5}$)&	3	&3&	1&	3	&2	&2&	1&	2&	13	&6&	2&3\\
X-WAM	&1	&4	&4	&4&	4	&4&	4	&4	&7	&4	&4&4\\
X-VLA&	5&	5	&5	&5&	5	&5&	13&	5&	5&	5	&1&5\\
GigaWorld-Policy-0	&6	&6	&6	&6&11	&6&	14	&6	&9&	2	&11&6\\
Xiaomi-Robotics-0	&7	&7&	7	&7&	9&	7	&10	&7	&15	&8	&7&7\\
AHA-WAM&	8	&9	&8&	8	&7	&12	&5&	8&	10	&13	&5&8\\
StarVLA-$\alpha$&	9	&12	&11	&9	&8&	9	&7&	9	&2	&7	&13&9\\
GalaxeaVLA (G0)	&11&	8	&9	&11	&5&	8&	6	&12	&11	&10	&10&10\\
$\pi_{0}$	&10&	11	&10	&10&	10&	11	&9	&10	&12	&12	&8&11\\
GR00T-N1.7&	12	&10&	13	&12&	14&	10	&11	&11	&3	&11	&12&12\\
InternVLA-A1	&13	&13&	14	&13	&14	&14&	16&	13	&4	&14	&9&13\\
Fast-WAM	&14	&14&	12	&14	&13	&13&	8	&14	&16&	9	&16&14\\
LDA-1B	&15	&15	&15&	15	&12	&15	&15	&15	&1	&15	&14&15\\
Spirit v1.5	&16	&16	&16	&16	&16&	16&	12	&16	&14	&16	&15&16\\
\bottomrule[1pt]
\end{tabular}%
\end{table}%

\begin{wrapfigure}{r}{0.68\linewidth}
    \vspace{-1.05em}
    \centering
    \includegraphics[width=1.0\linewidth]{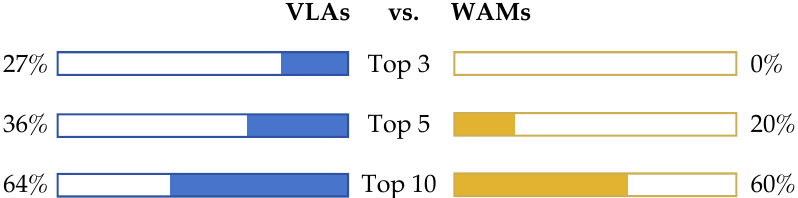}
    \caption{\textbf{Proportion of VLAs and WAMs ranked within the Top-3, Top-5, and Top-10 on RoboDojo-Sim.} Percentages are normalized by the total number of evaluated models in each paradigm.}
    \label{fig:q1_model_bubble_views}
    \vspace{-2em}
\end{wrapfigure}

Based on the overall rankings in Table~\ref{tab:comparison_dojo_sim1}, we further examine how the two model paradigms are distributed among the highest-ranked methods. Specifically, we compare the representation of VLAs and WAMs within the Top-3, Top-5, and Top-10 models, as summarized in Figure~\ref{fig:q1_model_bubble_views}. This rank-based view provides a more intuitive comparison of their competitiveness at different performance levels.
We can obtain the finding from Figure~\ref{fig:q1_model_bubble_views}:
\begin{enumerate}[label={}, leftmargin=10pt, itemindent=0pt, itemsep=0.3em, topsep=0.3em]
    \item[] \textit{\textbf{VLAs consistently demonstrate stronger overall performance than WAMs.}}  Across the Top-3, Top-5, and Top-10 rankings, VLAs exhibit a substantially higher representation.
\end{enumerate}

\subsection{Does a Larger Model Always Result in Better Performance?}
\begin{leaderbox}
\textcolor{boxline}{\textbf{Finding 2}: Larger model size does not guarantee stronger performance.}
\end{leaderbox}

\begin{figure}[ht]
\centering
\includegraphics[width=\linewidth]{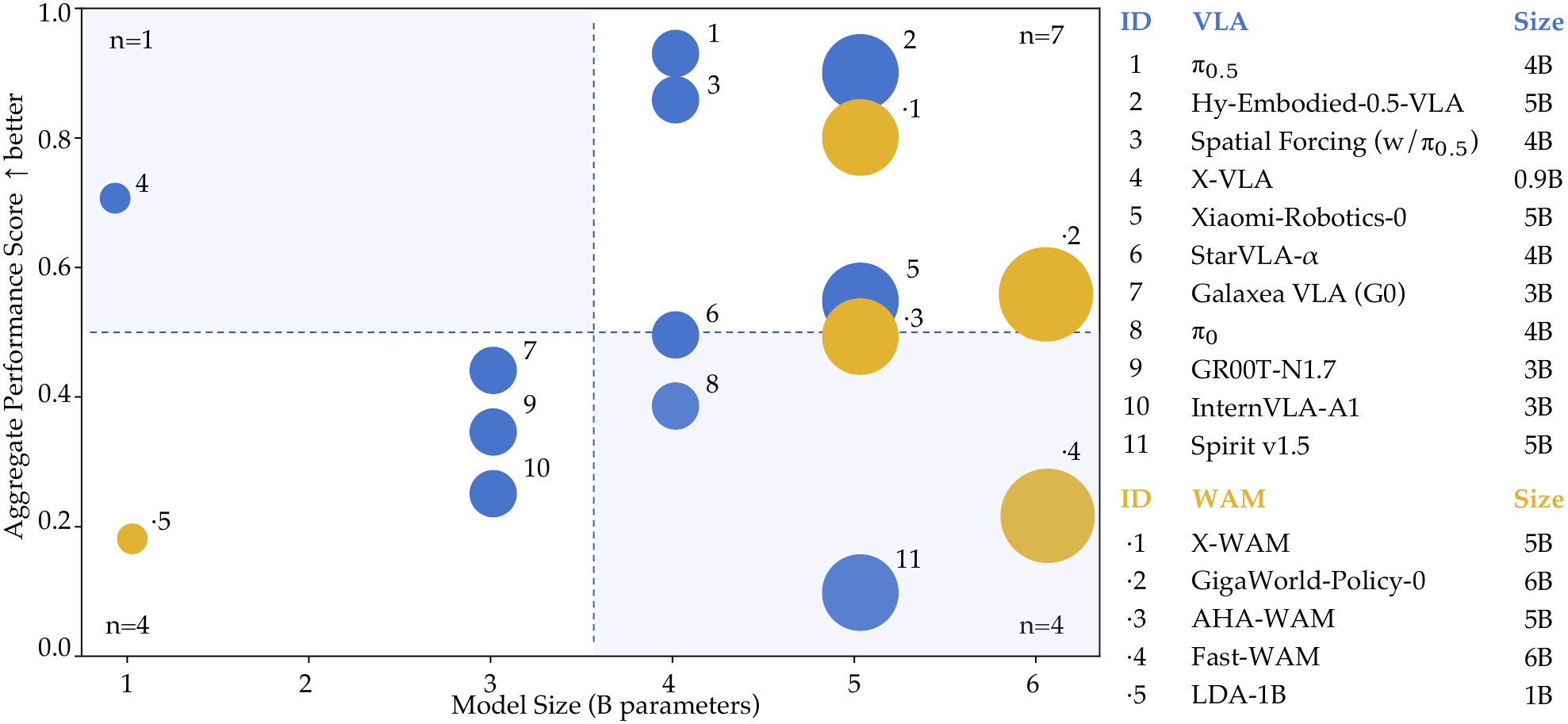}
\caption{\textbf{Model size versus overall ranking on RoboDojo-Sim.} Each point represents one model. The dashed lines indicate the median model size and median average rank across metrics, dividing the models into four regions. The shaded regions highlight models whose size and performance lie on opposite sides of the two medians, and $n$ denotes the number of models in each region.} 
\label{fig:q1_model_bubble_views1}
\end{figure}

Based on Table~\ref{tab:comparison_dojo_sim1}, we compare model size with the average ranking across the 11 evaluation metrics in Figure~\ref{fig:q1_model_bubble_views1}. The shaded regions capture two cases that do not follow a positive size--performance trend: models with relatively fewer parameters but stronger rankings, and models with relatively more parameters but weaker rankings. The results demonstrate that: 
\begin{enumerate}[label={}, leftmargin=10pt, itemindent=0pt, itemsep=0.3em, topsep=0.3em]
    \item[] \textit{\textbf{Larger model size does not necessarily translate into better embodied performance.}} The rankings show no clear positive correlation between parameter count and performance, with several relatively smaller models outperforming substantially larger counterparts.
\end{enumerate}

\subsection{Which model is generally the strongest?}
\begin{leaderbox}
\textcolor{boxline}{\textbf{Finding 3}: $\pi_{0.5}$ is generally the best model among different metrics. }
\end{leaderbox}

\begin{figure}[ht]
\centering
\includegraphics[width=0.85\linewidth]{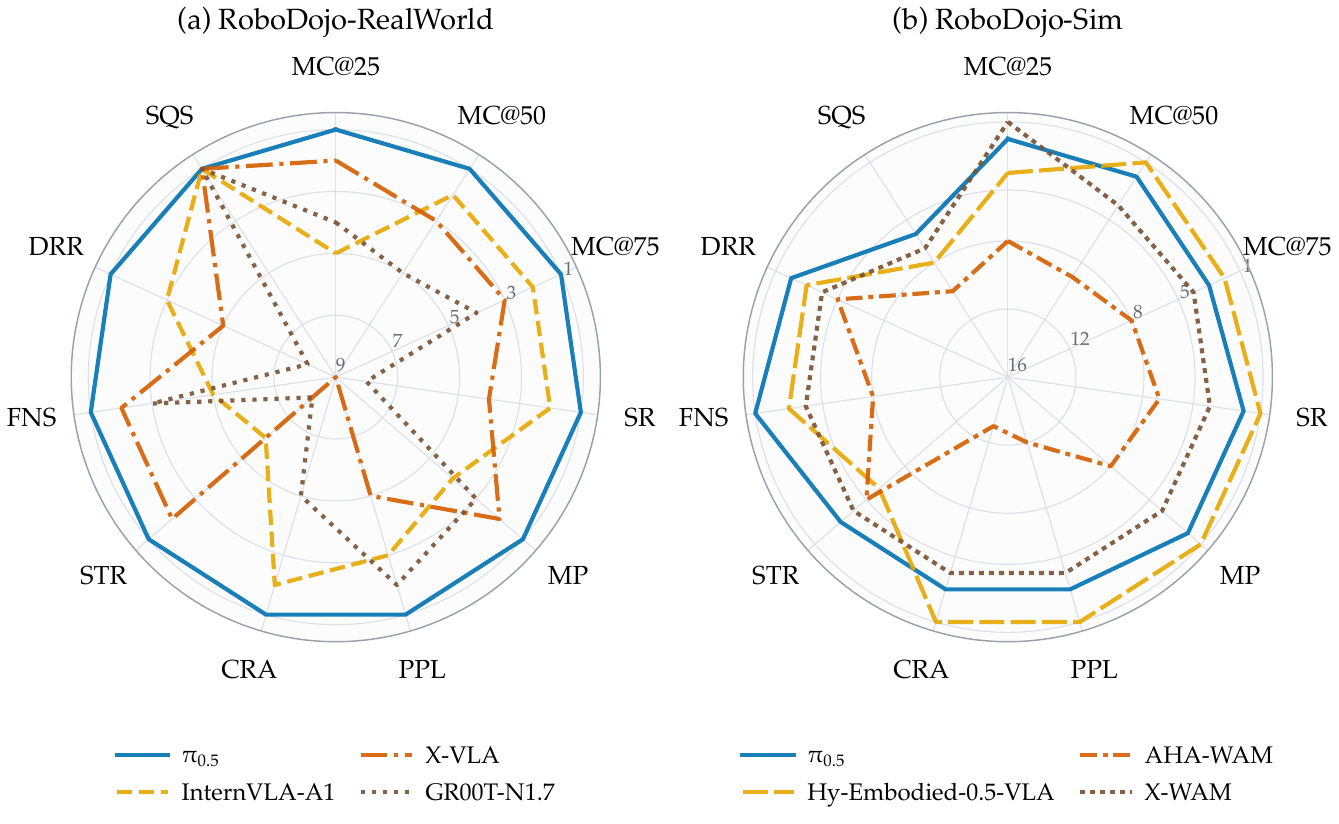}
\caption{\textbf{Performance comparison across different metrics in RoboDojo.} The further out the indicator is, the better the performance. SQS is not compared in the real-world setting because most models have low SR for a reliable estimate; its values are set to a common radius.} 
\label{fig:q2_rank_radar}
\end{figure}

As shown in Table \ref{tab:comparison_dojo_sim1} and Figure~\ref{fig:q2_rank_radar}, $\pi_{0.5}$ exhibits generally strong performance across multiple dimensions.

\subsection{What Tasks  Are Existing Embodied Models Relatively Better At?}
\begin{leaderbox}
\textcolor{boxline}{\textbf{Finding 4}: Existing embodied models are relatively better at performing \textit{Precision} tasks.}
\end{leaderbox}

\begin{figure}[ht]
\centering
\includegraphics[width=1.0\linewidth]{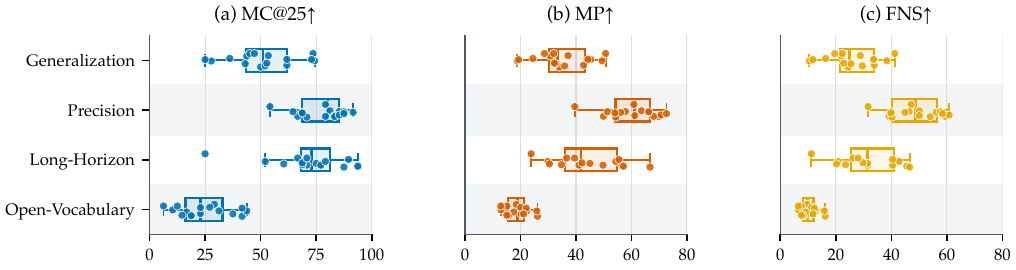}
\caption{\textbf{Task-category process profiles on RoboDojo-Sim.} Each point represents one model's category-level mean, with tasks weighted equally, and each box summarizes the distribution. The panels report metrics for 24 \textit{Generalization}, 8 \textit{Precision}, 8 \textit{Long-Horizon}, and 8 \textit{Open-Vocabulary} task conditions.}
\label{fig:q4_task_category_boxplots_horizontal}
\end{figure}

For assessing the difficulty of different tasks, we selected three metrics for analysis: MC@25, MP, and FNS. Based on Figure~\ref{fig:q4_task_category_boxplots_horizontal}, we obtained three conclusions:
\begin{enumerate}[label={}, leftmargin=10pt, itemindent=0pt, itemsep=0.3em, topsep=0.3em]
    \item[] \textit{\textbf{1) Precision tasks are currently the best-performing category, followed by Long-Horizon and Generalization tasks.}} \textit{Precision} tasks achieve relatively strong performance across MC@25, MP, and FNS. \textit{Long-Horizon} and \textit{Generalization} tasks form the next tier: models can often make meaningful early progress, but their progress depth and near-success behavior remain weaker than in \textit{Precision} tasks.
    \item[] \textit{\textbf{2) Open-vocabulary tasks are the most challenging category.}} Models achieve the lowest performance on this category, and their scores are tightly clustered at the low end. This indicates that models struggle more to make progress under open-vocabulary or less constrained task conditions.
    \item[] \textit{\textbf{3) Long-Horizon tasks is highly model-dependent.}} 
    Across all three metrics, \textit{Long-Horizon} tasks show the largest model performance variance. This larger spread makes \textit{Long-Horizon} tasks especially useful for distinguishing the relative capability of different embodied models.
\end{enumerate}

\vspace{-13pt}
\subsection{Does Simulation Performance Correlate with Real-World Performance?}

\begin{leaderbox}
\textcolor{boxline}{\textbf{Finding 5}: Simulation and real-world performance show only a weak positive correlation, with particularly poor correspondence on tasks requiring precise alignment and complex interaction.}
\end{leaderbox}

\begin{figure}[!ht]
\centering
\includegraphics[width=\linewidth]{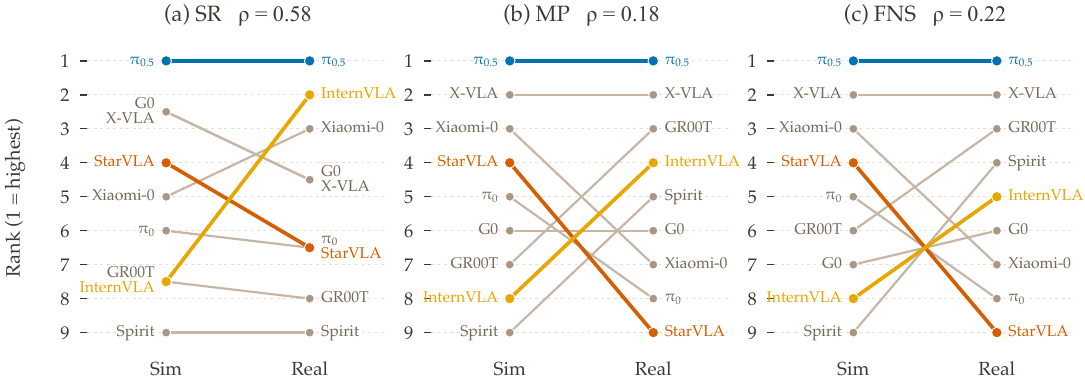}
\caption{\textbf{Simulation-to-real changes in performance ranking for the shared model set.} The three panels trace within-benchmark rank changes for SR, MP, and FNS across the nine shared models. Colors highlight $\pi_{0.5}$, InternVLA-A1, and StarVLA; all other models are shown in gray.}
\label{fig:q3_sim_to_real}
\end{figure}

\vspace{-5pt}
\begin{figure}[!h]
\centering
\includegraphics[width=0.8\linewidth]{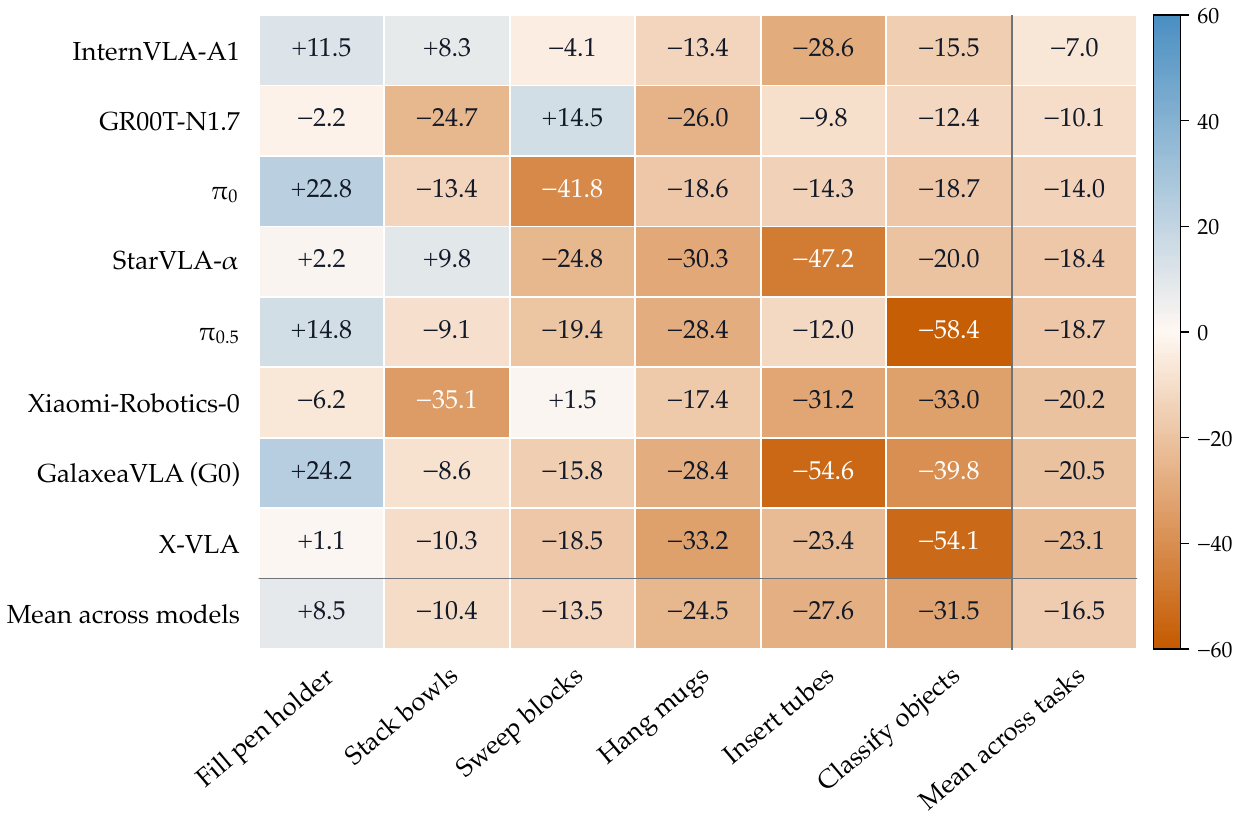}
\caption{\textbf{Simulation-to-real performance gaps in Max Progress (MP).} Each cell reports the Real-minus-Sim MP difference in percentage points across the six shared tasks. Orange cells (negative values) indicate lower progress in Real, whereas blue cells (positive values) indicate higher progress in Real.}
\label{fig:q3_hot}
\end{figure}

As shown in Figure~\ref{fig:q3_sim_to_real}, we select 9 models shared by the RoboDojo-Sim and RoboDojo-RealWorld leaderboards and analyze how their rankings change across the two settings.
Furthermore, in Figure~\ref{fig:q3_hot}, we restrict the analysis to 6 tasks shared between the simulation and real-world settings.
These tasks are designed to have comparable task objectives, making their metric values directly comparable.

Based on Figures~\ref{fig:q3_sim_to_real} and~\ref{fig:q3_hot}, we obtained three conclusions:
\begin{enumerate}[label={}, leftmargin=10pt, itemindent=0pt, itemsep=0.3em, topsep=0.3em]
    \item[] \textit{\textbf{1) Simulation performance cannot represent real-world performance.}} The ranking shift plots show that the correlations are mid-to-low (Spearman correlations \(\rho=0.18\)--\(0.58\)), and all models show negative Sim--Real performance gaps in the heatmap.
    \item[] \textit{\textbf{2) Tasks with higher spatial tolerance and simpler contact dynamics exhibit smaller Sim--Real gaps.}} \textit{Fill pen holder} shows a positive average gap, and tasks such as \textit{Stack bowls} or \textit{Sweep blocks} show milder degradation for some models, suggesting these simpler tasks are easier to transfer to real world.
    \item[] \textit{\textbf{3) Tasks requiring precise alignment or complex contact show substantially larger degradation under real-world deployment.}} Tasks such as \textit{Classify objects}, \textit{Hang mugs}, and \textit{Insert tubes} produce large negative gaps across most models, exposing real-world execution challenges that are less fully captured in the simulation environments.
\end{enumerate}
\vspace{-20pt}
\section{Conclusion}\label{sec:conclusion}
\setnavsection{sec:conclusion}

We presented PRM-as-a-Judge 1.5, a toolkit for robot process assessments, to enable robotic evaluations to move beyond binary success rates or rule-based scores
The 1.5 OPD metric suite further introduces three conditioned metrics based on the metrics introduced in PRM-as-a-Judge 1.0, making the OPD metric system more comprehensive.
Our large-scale assessments of mainstream embodied models show that process assessment can expose more detailed behavioral signatures.
We release the metric implementation and visualization tools for the community to use, to support a more open evaluation ecosystem for embodied intelligence.

\paragraph{Discussion \& Future Directions.}

\vspace{-2ex}
\begin{enumerate}[
    leftmargin=1.35em,
    itemsep=0.45em,
    topsep=0.35em,
    parsep=0em
]
    \item[] \textit{\textbf{1) Better PRMs.}} Our results suggest several directions for improving the progress judge itself.

    \vspace{-0.5ex}
    \begin{itemize}[
        leftmargin=1.2em,
        itemsep=0.2em,
        topsep=0.2em,
        parsep=0em
    ]
        \item \textbf{Temporal context.} Progress often depends on execution history. In repetitive or compositional tasks, the same local motion may indicate progress or regression depending on prior states. We therefore see sequence-style modeling as a more natural direction for future judges; offline assessment can further use both past and future observations for more stable judgments.

        \item \textbf{Negative-progress supervision.} Positive progress is easier to define from successful or expert trajectories, while regression is harder to supervise because the amount of progress lost after a failed action is often ambiguous. More real failed and regressive trajectories, together with better negative supervision, could improve Falling recognition.

        \item \textbf{Low-cost adaptation.} New tasks, viewpoints, embodiments, and corner cases will continue to appear. A practical base judge should therefore support low-cost adaptation to new evaluation settings without requiring full retraining.

        \item \textbf{Richer supervision.} Learning progress from rich multimodal observations with only scalar progress targets provides limited supervision for temporal dynamics. Future-state prediction, video prediction, and other temporal objectives could provide additional signals for learning task progress.
    \end{itemize}

    \item[] \textit{\textbf{2) Richer evidence for process diagnosis.}} Future assessment systems could combine progress curves with additional visual, state, contact, or semantic evidence, making the generated reports more interpretable and providing clearer explanations for observed execution patterns.

    \item[] \textit{\textbf{3) Closing the evaluation--data--training loop.}} Process assessment can already identify specific failure patterns and provide concrete directions for model improvement. These diagnoses can further guide failure-targeted data collection and training strategies, such as data reweighting and objective design, closing the loop between evaluation, data, and training.
\end{enumerate}
\section{Author List}
\label{sec:contributors}
\setnavsection{sec:contributors}

\setlength{\parskip}{0pt}

\begin{multicols}{2}

\begin{itemize}
    \setlength{\itemsep}{0.5em}
    \setlength{\parsep}{0pt}
    \setlength{\parskip}{0pt}
    \item Yuyang Liu$^{*}$
    \item Yanqing Shen$^{*}$
    \item Ruike Chen
    \item Jifan Zhao
    \item Yuxuan Tian
    \item Yichi Zhang
    \item Tianfeng Long
    \item Zixuan Yin
    \item Yipu Wang
    \item Ziheng Qin
    \item Wenxing Tan
    \item Yang Shi
    \item Mingyu Cao
    \item Runze Xiao
    \item Ziqi Wang
    \item Zhixin Yin
    \item Shiwei Chu
    \item Yi-Fan Zhang
    \item Yao Mu
    \columnbreak
    \item Yuheng Ji$^{\dagger}$
    \item Yihao Wang
    \item Jun Yan
    \item Zhongyuan Wang
    \item Pengwei Wang$^{\text{\Letter}}$
    \item Xiaolong Zheng$^{\text{\Letter}}$
\end{itemize}
\end{multicols}

\let\thefootnote\relax
\footnotemark\footnotetext{$^{*}$ Equal Contribution.}

\let\thefootnote\relax
\footnotetext{$^{\dagger}$ Project Lead.}

\let\thefootnote\relax
\footnotetext{$^{\text{\Letter}}$ Corresponding Authors: pwwang@baai.ac.cn, xiaolong.zheng@ia.ac.cn.}

\begingroup
\linespread{1}\selectfont
\renewcommand{\refname}{}
\section*{References}\phantomsection\label{sec:references}
\setnavsection{sec:references}
\vspace{-6ex}
\bibliographystyle{JudgeModel}

\bibliography{TR_Ref}
\endgroup

\newpage

\appendix

\section*{Appendix}\label{sec:appendix}
\setnavsection{sec:appendix}

\section{Related Work}
\label{app:related_work}

\subsection{Evaluation Metrics and Paradigms in Robotics}

Existing robot evaluation methods can be grouped by the information used to produce their scores. 

\textbf{Binary outcome metrics.} Binary Success Rate (SR) remains the standard metric for robotic manipulation \citep{song2026evaluation, vlaadapter, bai2025embodied, VLA-RFT}, but compresses an entire trajectory into a single success or failure label. It assigns the same score to a direct and smooth success as to one involving repeated attempts, pauses, or recovery. Likewise, a failure near the final step is treated the same as one with little task progress. SR therefore provides limited information about execution quality and failure causes \citep{wang2025roboeval, elmallah2025score, bai2025towards}. 
\textbf{Rule-based scoring.} Human evaluators or manually designed rules can assign scores to predefined stages and behaviors. This offers finer-grained feedback, but each task requires its own stages, thresholds, and weights; designing these rules is labor-intensive, and human judgments may vary \citep{orca, liu2026trustworthy,zhou2025autoeval, RoboBrain}.
\textbf{Auxiliary-state metrics.} Other methods evaluate intermediate progress, path length, smoothness, contact, or mechanical effort using object poses, joint states, and other environment signals \citep{anderson2018evaluation,zhu2020robosuite,gu2023maniskill2}. RoboEval, for example, combines stage completion with measures of efficiency, coordination, and stability \citep{wang2025roboeval}. These methods provide detailed diagnostics, but rely on privileged states and task-specific evaluation logic, which limits their use in real-world or black-box environments and increases the cost of supporting new tasks. 

\textit{\textbf{Our work.}} \textit{\textbf{PRM-as-a-Judge}} requires only a task description and trajectory video, and automatically derives fine-grained outcome, process, and diagnostic metrics. This simple input interface makes evaluation easier to deploy and scale across tasks, robot embodiments, simulators, and real-world systems.

\subsection{Process Reward Models in Robotics}

PRMs were originally developed to turn sparse task outcomes into dense step-level rewards for robot learning \citep{sermanet2018time, ma2022vip, ma2023liv, gilpin2024generative, chen2025sarm}. When applied throughout an episode, their predictions form a progress curve that captures advancement, stagnation, regression, and recovery, making them naturally suitable for process evaluation. Existing PRMs mainly differ in the temporal context used for prediction. \textbf{Pair-style PRMs} compare two observations~\citep{zhai2025vision}: Robo-Dopamine \citep{robodopamine} predicts the progress change between before-and-after states. \textbf{Sequence-style PRMs} jointly process multiple frames or longer video segments~\citep{liang2026robometer,chen2026topreward,liu2026passive,zhang2026recurrent}. Robometer~\citep{liang2026robometer} estimates frame-level progress and trajectory-level quality from complete trajectories, while WVM \citep{wang2026world} uses a pretrained video generation model (VGM) to capture visual dynamics and predict a sequence of progress values. Compared with conventional VLM-based PRMs, its VGM backbone provides stronger temporal representations and reduces reliance on progress supervision alone \citep{lv2026viva}.

\textit{\textbf{Our work.}} We evaluate representative models covering pair-style and sequence-style designs under a common evaluation protocol. We study the failure-type distribution, context-dependent task patterns, and computational efficiency, providing practical guidance for using PRMs as reliable robot judges.
\section{Metric Definitions}
\label{app:metric}

\subsection{Progress Curve Construction}

For each rollout, the judge outputs are converted into normalized task progress and aligned with the sampled video time steps. The resulting progress curve is written as
\begin{equation}
    p_{0:T} = (p_0,\ldots,p_T),
    \qquad
    p_t \in [0,1],
\end{equation}
where $p_t$ denotes the task progress at time step $t$. A value closer to $1$ indicates that the rollout is closer to task completion. This progress curve provides the common input for all OPD metrics. Before computing the metrics, we apply Gaussian smoothing to reduce local prediction noise.

\subsection{OPD Metrics}

PRM-as-a-Judge organizes curve-based evaluation into three complementary levels. The \textit{Outcome} level summarizes task reachability, the \textit{Process} level evaluates how efficiently task progress is achieved over the full rollout, and the \textit{Diagnosis} level characterizes execution patterns such as regression, stagnation, near-success failure, recovery, and success quality.

Each metric is first computed for each eligible rollout. At the task and model levels, continuous metrics are reported as the median across eligible rollouts; specifically, FNS, DRR, and SQS are aggregated only over failed rollouts, rollouts that experience a drawdown, and successful rollouts, respectively. MC@$q$ and SR are reported as the percentage of rollouts that reach the corresponding milestone or complete the task.

\paragraph{Outcome Level.}
Outcome metrics describe how far a rollout progresses toward task completion.

\begin{itemize}

\item  \textbf{Max Progress (MP).}
MP is the highest progress reached during the rollout:
\begin{equation}
    \mathrm{MP}
    =
    \max_{0\leq t\leq T} p_t.
\end{equation}
A higher MP indicates that the rollout reaches a later task stage.

\item \textbf{Milestone Coverage (MC@$q$).}
MC@$q$ indicates whether the rollout reaches a predefined progress milestone. For $q\in\{25,50,75,100\}$,
\begin{equation}
    \mathrm{MC@}q
    =
    \mathbb{I}
    \left[
        \mathrm{MP}\geq\frac{q}{100}
    \right].
\end{equation}
When averaged over rollouts, MC@$q$ is the proportion of rollouts that reach milestone $q$. Higher values indicate stronger reachability at the corresponding task stage.
\end{itemize}

\paragraph{Process Level.}
Process metrics describe how progress is accumulated along the execution path.

\begin{itemize}

\item \textbf{Path-weighted Progress Length (PPL).}
PPL measures path efficiency over the full rollout, weighted by the highest progress reached:
\begin{equation}
    \mathrm{PPL}
    =
    \mathrm{MP}
    \cdot
    \frac{\mathrm{MP}}
    {\sum_{t=1}^{T}\left|p_t-p_{t-1}\right|}.
\end{equation}
PPL is set to $0$ when the denominator is $0$. The first $\mathrm{MP}$ weights the score by the highest progress reached, while the second term measures path efficiency by comparing the required progress with the actual length of the full progress path. A higher PPL indicates that the rollout achieves greater progress more efficiently, with less backtracking or repeated correction.

\end{itemize}

\paragraph{Diagnosis Level.}
Diagnosis metrics characterize the execution patterns behind failed, unstable, recovering, and successful rollouts.

\begin{itemize}

\item  \textbf{Cumulative Regret Area (CRA).}
CRA measures how far and how long the rollout remains below its best-so-far progress:
\begin{equation}
    \mathrm{CRA}
    =
    \frac{1}{(T+1)\,\mathrm{MP}}
    \sum_{t=0}^{T}
    \left(
        \max_{0\leq i\leq t}p_i-p_t
    \right).
\end{equation}
A higher CRA indicates more severe or persistent regression during execution.

\item \textbf{Stagnation Ratio (STR).}
STR measures the proportion of consecutive time steps with negligible progress change:
\begin{equation}
    \mathrm{STR}
    =
    \frac{1}{T}
    \sum_{t=1}^{T}
    \mathbb{I}
    \left[
        \left|p_t-p_{t-1}\right|<\epsilon
    \right],
\end{equation}
where $\epsilon>0$ is the threshold for negligible progress change. A higher STR indicates more frequent stagnation or hesitation.

\item \textbf{Failure Near-Success (FNS).}
For failed rollouts, FNS measures how close the execution comes to task completion:
\begin{equation}
    \mathrm{FNS}
    =
    0.5\,\mathrm{MP}
    +0.3\,\mathrm{MC@75}
    +0.2\,\mathrm{MC@50}.
\end{equation}
A higher FNS indicates that the rollout fails after reaching more meaningful mid-to-late task stages.

\item \textbf{Drawdown Recovery Ratio (DRR).}
For rollouts that experience a drawdown, DRR measures how much the execution recovers after its largest progress loss. The drawdown at time step $t$ is $d_t=\max_{0\leq i\leq t}p_i-p_t$, and $t^\star\in\arg\max_t d_t$ denotes the point of the largest drawdown. DRR is defined as
\begin{equation}
    \mathrm{DRR}
    =
    \min\left(
        1,
        \frac{
            \max_{t^\star\leq t\leq T}p_t-p_{t^\star}
        }{
            d_{t^\star}
        }
    \right),
    \qquad d_{t^\star}>0.
\end{equation}
The numerator measures the best subsequent recovery, while the denominator is the size of the largest progress loss. A higher DRR indicates stronger recovery, with $1$ representing full recovery. Rollouts without a drawdown are excluded from DRR aggregation.

\item  \textbf{Success Quality Score (SQS).}
For successful rollouts, SQS summarizes path efficiency, regression, and stagnation:
\begin{equation}
    \mathrm{SQS}
    =
    0.5\,\mathrm{PPL}
    +0.3\left(1-\mathrm{CRA}\right)
    +0.2\left(1-\mathrm{STR}\right).
\end{equation}
A higher SQS indicates that the task is completed efficiently, with less regression and stagnation.

\end{itemize}
\section{RoboPulse++: Benchmarking Progress Judge Models}
\label{app:benchmark}

RoboPulse++ is an interval-level benchmark for evaluating progress judge models across diverse robot embodiments, environments, tasks, and execution dynamics. Each annotated interval is labeled as \textit{Rising} or \textit{Falling}, providing a common target for evaluating model predictions throughout robot execution. This appendix presents the dataset composition and annotation protocol, evaluation procedures for specialized PRMs and general-purpose VLMs, and their progress-assessment accuracy.

\subsection{From Pairwise Comparison to Interval-Level Progress Assessment}

RoboPulse~\citep{PRM-as-a-Judge} evaluates relative progress between two states from the same trajectory. Then RoboPulse++ extends this formulation to temporal intervals, where each interval is annotated with its progress direction. This supports process-judge evaluation throughout complete robot trajectories.

We use directional interval labels because absolute completion percentages lack a consistent ground truth across diverse robot tasks and execution processes. When an execution involves partial completion, regression, or recovery, assigning an exact progress value requires mapping task states to a shared numerical scale. The direction of progress within a temporal interval can be identified more directly from task-relevant state changes. Each interval is therefore annotated as \textit{Rising} or \textit{Falling}.

\vspace{-1em}
\subsection{Dataset Construction and Composition}

\begin{wrapfigure}{r}{0.34\linewidth}
    \vspace{-3em}
    \centering
    \includegraphics[width=1.0\linewidth]{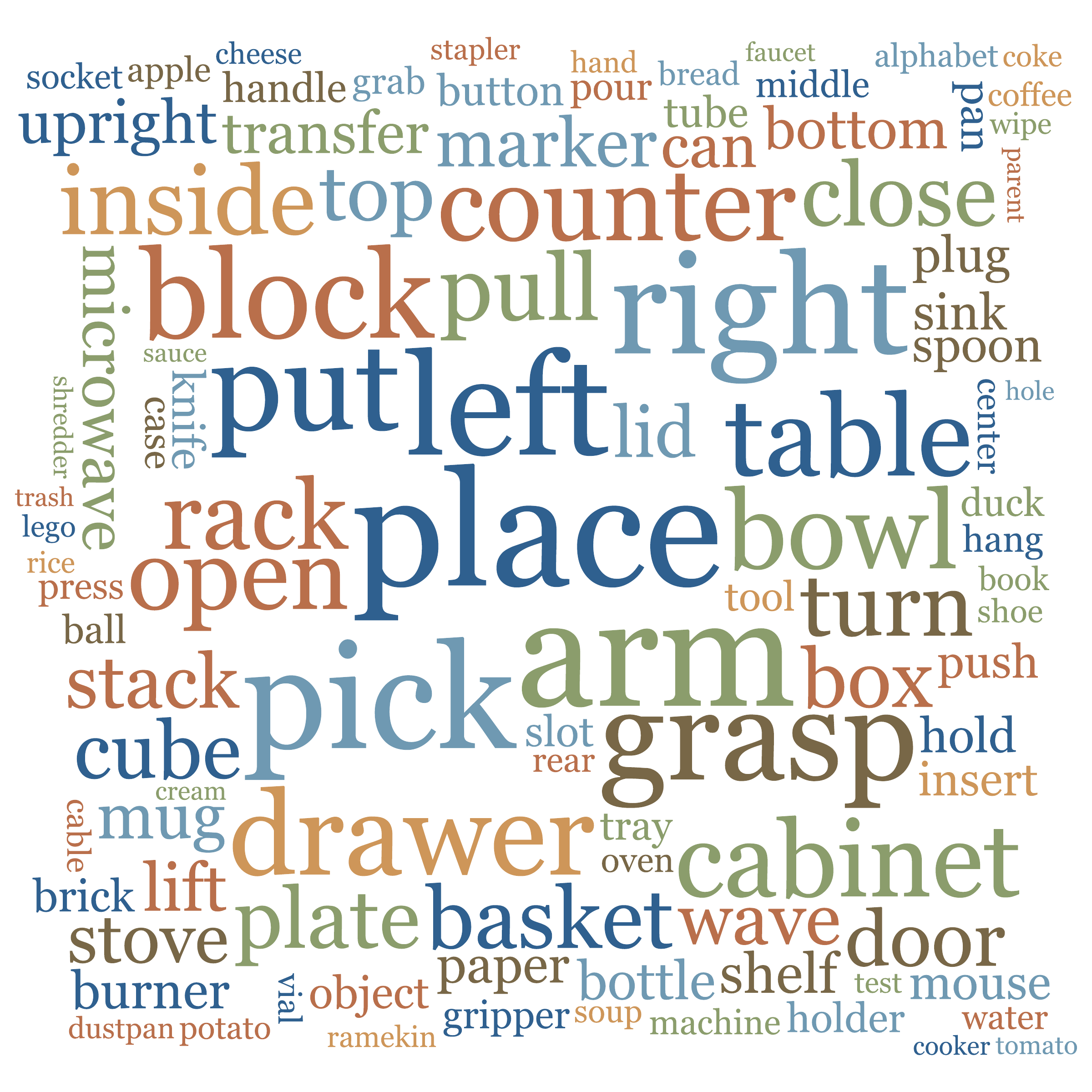}
    \vspace{-1em}
    \caption{\textbf{Task-semantic coverage of RoboPulse++.}}
    \label{fig:robopulse_task_wordcloud}
\end{wrapfigure}

RoboPulse++ contains 700 manipulation trajectories spanning 275 task entries, 17,052 frames, and 2,244 human-annotated intervals across diverse real-world and simulation sources.
The benchmark captures task-relevant progress and regression throughout robot execution.
Table~\ref{tab:robopulse_statistics} summarizes its composition.

\textbf{\textit{Data coverage.}}
The dataset includes 439 real-world trajectories (62.7\%) and 261 simulation trajectories (37.3\%).
These sources cover diverse robot embodiments, scenes, camera viewpoints, and execution conditions.

\textbf{\textit{Task coverage.}}
RoboPulse++ spans atomic manipulation skills, compositional tasks, and long-horizon execution.
The tasks include grasping, placing, pushing, pulling, articulated-object manipulation, object sorting and stacking, tool use, and multi-stage manipulation. Figure~\ref{fig:robopulse_task_wordcloud} summarizes the semantic coverage of the task descriptions.

\begin{table}[ht]
\small
    \centering
    \caption{\textbf{Data statistics of RoboPulse++ across different sources.}}
    \label{tab:robopulse_statistics}
    \begin{tabular}{@{}lcccc@{}}
        \toprule
        \textit{\textbf{{Data source}}} & {Episodes} & {Tasks} & {Frames} & {Intervals} \\
        \midrule
        \rowcolor[rgb]{.949,.949,.949}
        \multicolumn{5}{l}{\textit{\textbf{Real-World Data}}} \\
        RoboChallenge-Table30~\citep{robochallenge}    & 79  & 25  & 3,448  & 882 \\
        RoboChallenge-Table30-V2~\citep{robochallenge} & 79  & 26  & 2,068  & 169 \\
        ViFailback~\citep{ViFailback}                  & 265 & 103 & 4,760  & 544 \\
        RoboFAC-Real~\citep{robofac}                   & 16  & 6   & 261    & 33  \\
        \rowcolor[rgb]{.949,.949,.949}
        \multicolumn{5}{l}{\textit{\textbf{Simulation Data}}} \\
        LIBERO~\citep{libero}                          & 111 & 39  & 2,462  & 239 \\
        RoboCasa~\citep{robocasa}                      & 34  & 24  & 1,935  & 79  \\
        SimplerEnv~\citep{li2024simplerenv}            & 24  & 9   & 566    & 104 \\
        RoboTwin 2.0~\citep{robotwin20}                & 22  & 7   & 822    & 58  \\
        RoboFAC-Sim~\citep{robofac}                    & 70  & 36  & 730    & 136 \\
        \midrule
        \textbf{Total} & \textbf{700} & \textbf{275} & \textbf{17,052} & \textbf{2,244} \\
        \bottomrule
    \end{tabular}
\end{table}

\subsection{Interval Annotation Protocol}

For each trajectory, annotators first review the complete execution together with its natural-language task instruction, and then divide the trajectory into temporally contiguous intervals $[t_s,t_e]$ according to the evolution of task progress.
Each interval is assigned a label $y_{[t_s,t_e]}\in\{-1,+1\}$ based on the task-relevant state changes:
\begin{itemize}[leftmargin=2em]
    \item \textbf{\textit{Rising} ($+1$):} \ 
    Positive progress toward the task goal, such as moving an object toward its target state, bringing the gripper closer to the target object, or completing a required manipulation step.

    \item \textbf{\textit{Falling} ($-1$):} \ 
    Negative progress, where the task-relevant state moves away from the goal, such as moving an object away from its target state or reversing previously achieved task progress.
\end{itemize}

Annotators use the full trajectory as temporal context when determining interval labels.
The annotation interface allows them to scrub through the video and directly mark the boundaries of each progress interval.
This context helps distinguish transient local fluctuations from meaningful progress or regression, especially when changes between nearby observations are subtle.

\subsection{Evaluation Protocol}
\label{sec:robopulse_eval}

RoboPulse++ evaluates progress judge models by comparing their predictions with the human-annotated progress direction within each interval. The target space consists of \textit{Rising} ($+1$) and \textit{Falling} ($-1$).

We evaluate specialized PRMs and general-purpose VLMs under a common protocol. We use \textit{Pair Style} for progress judgment from adjacent observations and \textit{Sequence Style} for progress judgment from a temporally ordered observation sequence. General-purpose VLMs provide a reference for how broadly pretrained multimodal models perform on the same progress-judgment task without task-specific reward modeling. All model outputs are converted to the same \textit{Rising}/\textit{Falling} prediction space. All videos are uniformly sampled at 1 FPS and paired with their natural-language task instructions.

\subsubsection{PRM Evaluation}

\textbf{Baselines.} We evaluate Robo-Dopamine~\citep{robodopamine}, RoboMeter~\citep{liang2026robometer}, LRM~\citep{LRM}, TOPReward~\citep{chen2026topreward}, VLAC~\citep{vlac}, GVL~\citep{gvl}, and PRIMO~\citep{PRIMO}. For models with multiple inference formulations, we evaluate Robo-Dopamine with Forward, Backward, and Incremental inference, and LRM with Temporal and Absolute inference. Together, these baselines cover both \textit{Pair Style} and \textit{Sequence Style}.

\textbf{Output Unification.} Different PRMs produce different native outputs, including absolute progress scores, local progress increments, and categorical progress directions. For an absolute progress prediction $p_t$, we compute the local change as $d_t = p_t - p_{t-1}$. Native progress increments are directly used as $d_t$, while categorical predictions are mapped directly to \textit{Rising} ($+1$) or \textit{Falling} ($-1$).

\textbf{Temporal Smoothing.} For continuous PRM outputs, we apply a five-frame causal moving average to $d_t$ before determining its direction. The sign of the smoothed change gives the final progress prediction: positive and negative changes correspond to \textit{Rising} and \textit{Falling}, respectively. Categorical outputs are used directly without smoothing.

\subsubsection{General-Purpose VLM Evaluation}

\textbf{Baselines.} We evaluate GPT-5.4~\citep{GPT54}, Gemini 3.1 Pro~\citep{Gemini-3.1-pro}, Qwen 3.6 Plus~\citep{qwen36_35b_a3b}, and Claude Sonnet 4.6~\citep{sonnet46}, each under both Pair Style and Sequence Style.

\paragraph{Pair Style.}
For each adjacent observation pair $(o_{t-1}, o_t)$, the VLM directly predicts the progress direction as $\hat{y}_t^{\mathrm{pair}} \in \{-1,+1\}$.

\paragraph{Sequence Style.}
For each human-annotated interval, the VLM receives all sampled observations in temporal order and predicts an absolute task-progress score $p_t \in [0,100]$ for each observation, where 0 and 100 correspond to no achieved task progress and task completion, respectively. The directional prediction is then obtained from consecutive scores as $\hat{y}_t^{\mathrm{seq}} = \operatorname{sign}(p_t - p_{t-1})$.

\subsubsection{Metric Computation}

\textbf{Label Matching.} Each annotated interval $[t_s,t_e]$ has a single ground-truth progress label $y_{[t_s,t_e]} \in \{-1,+1\}$, corresponding to \textit{Rising} or \textit{Falling}. Judge predictions within the interval are evaluated against this label.

\textbf{Boundary Filtering.} To reduce ambiguity around transitions between progress states, we exclude predictions from the first two and last two sampled time steps of each annotated interval.

\textbf{Metrics.} We report Macro-F1 and Accuracy, together with class-wise Precision, Recall, and F1 for \textit{Rising} and \textit{Falling}. Metrics are computed over all valid predictions pooled across episodes.

\subsection{Experimental Results}
Table~\ref{tab:robopulse_main} summarizes the progress-direction results across specialized PRMs and general-purpose VLMs. 

\begin{table*}[ht]
    \centering
    \small
    \setlength{\tabcolsep}{3.5pt}  
    \caption{\textbf{Progress-direction performance of judge models on RoboPulse++ (700 episodes).}}
    \label{tab:robopulse_main}
    \begin{tabular}{@{}lcccccccc@{}}
        \toprule
        \multirow{2}{*}{\textbf{Judge Model}} & \multicolumn{3}{c}{\textit{\textbf{Rising}}} & \multicolumn{3}{c}{\textit{\textbf{Falling}}} & \multicolumn{2}{c}{\textit{\textbf{Overall}}} \\
        \cmidrule(lr){2-4}\cmidrule(lr){5-7}\cmidrule(lr){8-9}
        & \textbf{Precision} & \textbf{Recall} & \textbf{F1 Score} & \textbf{Precision} & \textbf{Recall} & \textbf{F1 Score} & \textbf{Macro-F1} & \textbf{Accuracy} \\
        \midrule
        \rowcolor[rgb]{.949,.949,.949}
        \multicolumn{9}{l}{\textbf{\textit{Specialized PRMs}}} \\
        Robo-Dopamine (Forward)     & 0.92 & \textbf{0.92} & \textbf{0.92} & \textbf{0.79} & 0.50 & 0.61 & \textbf{0.77} & \textbf{0.84} \\
        Robo-Dopamine (Backward)    & 0.92 & 0.88 & 0.90 & 0.72 & 0.55 & \textbf{0.63} & 0.76 & 0.82 \\
        Robo-Dopamine (Incremental) & \textbf{0.93} & 0.84 & 0.88 & 0.62 & \textbf{0.57} & 0.59 & 0.74 & 0.79 \\
        RoboMeter                   & 0.89 & 0.86 & 0.87 & 0.47 & 0.54 & 0.50 & 0.69 & 0.80 \\
        LRM (Temporal)              & 0.84 & 0.71 & 0.77 & 0.29 & 0.42 & 0.34 & 0.56 & 0.66 \\
        LRM (Absolute)              & 0.90 & 0.31 & 0.46 & 0.35 & 0.21 & 0.26 & 0.36 & 0.29 \\
        TOPReward                   & 0.85 & 0.75 & 0.80 & 0.28 & 0.38 & 0.32 & 0.56 & 0.68 \\
        VLAC                        & 0.89 & 0.38 & 0.54 & 0.26 & 0.22 & 0.24 & 0.39 & 0.35 \\
        GVL                         & 0.90 & 0.78 & 0.84 & 0.51 & 0.47 & 0.49 & 0.66 & 0.72 \\
        PRIMO                       & 0.86 & 0.75 & 0.80 & 0.20 & 0.05 & 0.08 & 0.44 & 0.62 \\
        \midrule
        \rowcolor[rgb]{.949,.949,.949}
        \multicolumn{9}{l}{\textbf{\textit{General-Purpose VLMs -- Pair Style}}} \\
        GPT-5.4                     & \textbf{0.94} & 0.37 & 0.53 & 0.47 & 0.15 & 0.23 & 0.38 & 0.33 \\
        Gemini 3.1 Pro              & 0.91 & \textbf{0.71} & \textbf{0.80} & 0.50 & \textbf{0.29} & \textbf{0.37} & \textbf{0.58} & \textbf{0.63} \\
        Qwen 3.6 Plus               & 0.91 & 0.68 & 0.78 & \textbf{0.51} & 0.21 & 0.29 & 0.54 & 0.59 \\
        Claude Sonnet 4.6           & 0.93 & 0.38 & 0.54 & 0.34 & 0.21 & 0.26 & 0.40 & 0.35 \\
        \midrule
        \rowcolor[rgb]{.949,.949,.949}
        \multicolumn{9}{l}{\textbf{\textit{General-Purpose VLMs -- Sequence Style}}} \\
        GPT-5.4                     & 0.96 & 0.59 & 0.73 & 0.79 & 0.14 & 0.24 & 0.49 & 0.50 \\
        Gemini 3.1 Pro              & 0.90 & 0.42 & 0.58 & 0.81 & 0.07 & 0.12 & 0.35 & 0.36 \\
        Qwen 3.6 Plus               & \textbf{0.97} & \textbf{0.61} & \textbf{0.75} & \textbf{0.87} & 0.13 & 0.23 & \textbf{0.49} & \textbf{0.52} \\
        Claude Sonnet 4.6           & 0.95 & 0.54 & 0.69 & 0.82 & \textbf{0.16} & \textbf{0.26} & 0.48 & 0.46 \\
        \bottomrule
    \end{tabular}
\end{table*}

Two main patterns emerge:
\begin{enumerate}[
    leftmargin=1.35em,
    itemsep=0.35em,
    topsep=0.35em,
    parsep=0em
]
\item[] \textbf{\textit{1) Specialized PRMs provide the strongest and most balanced progress judgments.}} Robo-Dopamine (Forward) achieves the best overall Macro-F1 and Accuracy, substantially exceeding the most competitive general-purpose VLM setting, Gemini 3.1 Pro in Pair Style.

\item[]  \textbf{\textit{2) Falling progress remains materially harder to recognize than Rising progress.}} The best Falling F1 is 0.63, whereas the best Rising F1 is 0.92. This gap is driven primarily by lower Falling recall.
Robust detection of regression remains the principal limitation of the evaluated judges.
\end{enumerate}

\subsubsection{Falling Error Analysis}

We sampled 155 intervals labeled as \textit{Falling} and summarized two representative failure types: \textit{Interaction-relation failure} refers to regression during robot--object interaction, such as dropping a grasped object or failing to place it at the target location.
\textit{Task-order failure} refers to violations of the expected execution order, such as performing subtasks in an incorrect sequence or executing actions unrelated to the goal.

\begin{table}[ht]
    \centering
    \setlength{\tabcolsep}{10pt}
    \caption{\textbf{Failure-type distribution of \textit{Falling} intervals.}}
    \label{tab:robopulse_falling_taxonomy}
    \begin{tabular}{@{}lcc@{}}
        \toprule
        \textbf{Failure type} & \textbf{\# Intervals} & \textbf{Proportion} \\
        \midrule
        Interaction-relation failure & 121 & 78.1\% \\
        Task-order failure & 34 & 21.9\% \\
        \bottomrule
    \end{tabular}
\end{table}

As shown in Table~\ref{tab:robopulse_falling_taxonomy}, the error distribution is strongly concentrated in interaction-relation failures.
This result indicates that achieving precise real-world interaction is more difficult for current embodied models than understanding the high-level task logic.

\subsubsection{Context-Dependent Task Analysis}

Some progress changes depend on execution history. For example, a rollout may complete an intermediate subgoal and later undo it. In such cases, judging only a pair of states can miss the loss of previously achieved progress, while an ordered sequence provides the earlier states needed to recognize the regression. To examine this pattern, we compare Robo-Dopamine (Forward), a representative Pair Style PRM, with RoboMeter, a representative Sequence Style PRM, on context-dependent cases in RoboPulse++. Table~\ref{tab:robopulse_context_accuracy} reports their class-conditioned and overall accuracy.

\begin{table}[ht]
    \centering
    \setlength{\tabcolsep}{8pt}
    \caption{\textbf{Accuracy of representative Pair Style and Sequence Style PRMs on context-dependent task patterns.} \textit{Rising} and \textit{Falling} report class-conditioned accuracy, while Overall reports accuracy over all evaluated points.}
    \label{tab:robopulse_context_accuracy}
    \begin{tabular}{@{}lcc@{}}
        \toprule
        \textbf{Evaluation subset}
        & \textbf{Robo-Dopamine (Forward)}
        & \textbf{RoboMeter} \\
        \midrule
        \textit{Rising}  & 0.844 & \textbf{0.951} \\
        \textit{Falling} & 0.408 & \textbf{0.901} \\
        \midrule
        \textbf{Overall} & 0.746 & \textbf{0.940} \\
        \bottomrule
    \end{tabular}
\end{table}

RoboMeter performs better on these context-dependent cases, with the largest gap on \textit{Falling} (0.901 vs.\ 0.408). This pattern suggests that sequence-level context is particularly useful for recognizing regressions whose meaning depends on previously achieved states or subgoals.

\vspace{-5pt}
\section{Computational Cost of Progress Judges}
\label{app:computational_cost}

Large-scale process assessment can involve hundreds of rollout videos, making computational cost an important consideration when deploying PRMs. We therefore profile representative PRMs with different model scales and inference formulations to provide references for different computational budgets.

\textbf{Profiling Setup.} We evaluate the selected judges on the same set of 100 videos using a single NVIDIA H100 80 GB GPU. Each video has a resolution of $256\times256$ and an average duration of 75 seconds, amounting to approximately 2.1 hours of raw rollout video in total. All videos are uniformly sampled at 1 FPS. Table~\ref{tab:judge_efficiency} reports peak GPU memory and runtime at batch size 1.

\begin{table*}[ht]
\centering
\small
\setlength{\tabcolsep}{7pt}
\caption{\textbf{Computational cost of representative progress judge models.} }
\label{tab:judge_efficiency}
\begin{tabular}{lccc}
\toprule
{\textbf{\textit{Judge Model}}} &
{\textbf{Params. (B)}} &
\textbf{Peak Memory (GiB)} & \textbf{Runtime}\\
\midrule
Robo-Dopamine & 4 & 11 & 29.3 min \\ 
Robo-Dopamine & 8 & 19 & 31.7 min \\ 
VLAC & 2 & 6 & 11.3 min \\ 
RoboMeter & 4 & 12 & 46 s \\
TOPReward & 8 & 17 & 12.3 min \\
PRIMO (w/ CoT) & 7 & 17 & 19.6 h  \\
\bottomrule
\end{tabular}
\end{table*}

\paragraph{Efficiency Analysis.} At batch size 1, peak GPU memory ranges from 6 to 19 GiB. With sufficient GPU memory, batched inference can further improve the throughput of these judges. For example, on a single NVIDIA H100 80 GB GPU, increasing the batch size reduces the total runtime of Robo-Dopamine-4B from 29.3 to 7.8 minutes and Robo-Dopamine-8B from 31.7 to 9.1 minutes.

\newcommand{\best}[1]{\textbf{#1}}
\newcommand{\second}[1]{\underline{#1}}

\vspace{-10pt}
\section{Recovering Success Rate from Progress Curves}
\label{app:robodojo}

PRM-based evaluation derives fine-grained metrics from dense progress curves. We further examine whether the same progress curves can recover conventional model-level success rates on RoboDojo.

\subsection{Evaluation Setup}

RoboDojo~\citep{robodojo} contains 42 simulation tasks and 18 real-world tasks spanning diverse manipulation capabilities and embodiments. We use its released rollout videos for the model assessment. The alignment covers 6,076 rollouts, including 4,606 simulation and 1,470 real-world rollouts.

We use Robo-Dopamine (Forward), the default PRM in our main evaluation. 
A rollout is counted as successful when its maximum predicted progress reaches 1 ($\mathrm{MP}=1$), and the PRM-derived SR is computed as the proportion of successful rollouts.

\subsection{Success-Rate Consistency}
We compare the PRM-derived SR with the benchmark-reported RoboDojo SR at the model level. We report the mean absolute error (MAE), signed mean difference (defined as PRM-derived SR minus RoboDojo SR), and Spearman rank correlation across models. 
Table~\ref{tab:app_sr_alignment} summarizes the results.

\begin{table}[h]
\centering
\setlength{\tabcolsep}{10pt}
\caption{\textbf{Model-level consistency between PRM-derived and benchmark-reported RoboDojo SR.} }
\label{tab:app_sr_alignment}
\begin{tabular}{@{}lcccc@{}}
\toprule
\textbf{Setting} & \textbf{\# Models} & \textbf{SR MAE (pp)} & \textbf{Mean Diff. (pp)} & \textbf{Spearman $\rho$} \\
\midrule
Simulation & 16 & 1.57\% & +1.29\% & 0.88 \\
Real-World & 9 & 1.32\% & -0.35\% & 0.96 \\
\bottomrule
\end{tabular}
\begin{tablenotes}
\footnotesize
\item \textit{Note.} For Simulation, the benchmark-reported RoboDojo SR is recomputed over the four task categories used in this analysis with task-count weights $12{:}8{:}8{:}8$. For Real-World, the benchmark-reported RoboDojo SR is used directly because the three embodiment groups contain equal numbers of trials. PRM-derived SR is computed over the available rollout cases.
\end{tablenotes}
\end{table}

The PRM-derived SR is close to the benchmark-reported SR, with MAEs of 1.57 and 1.32 percentage points in Simulation and Real-World, respectively, and Spearman correlations of 0.88 and 0.96. This indicates that the progress curves retain conventional outcome-level information while supporting fine-grained process assessment.
\clearpage
\section{Visualization and Case Studies}
\label{app:more}
\subsection{Case Studies}
\label{sec:failure-pattern-visualization}

\begin{figure}[!b]
    \centering
    \includegraphics[width=\linewidth]{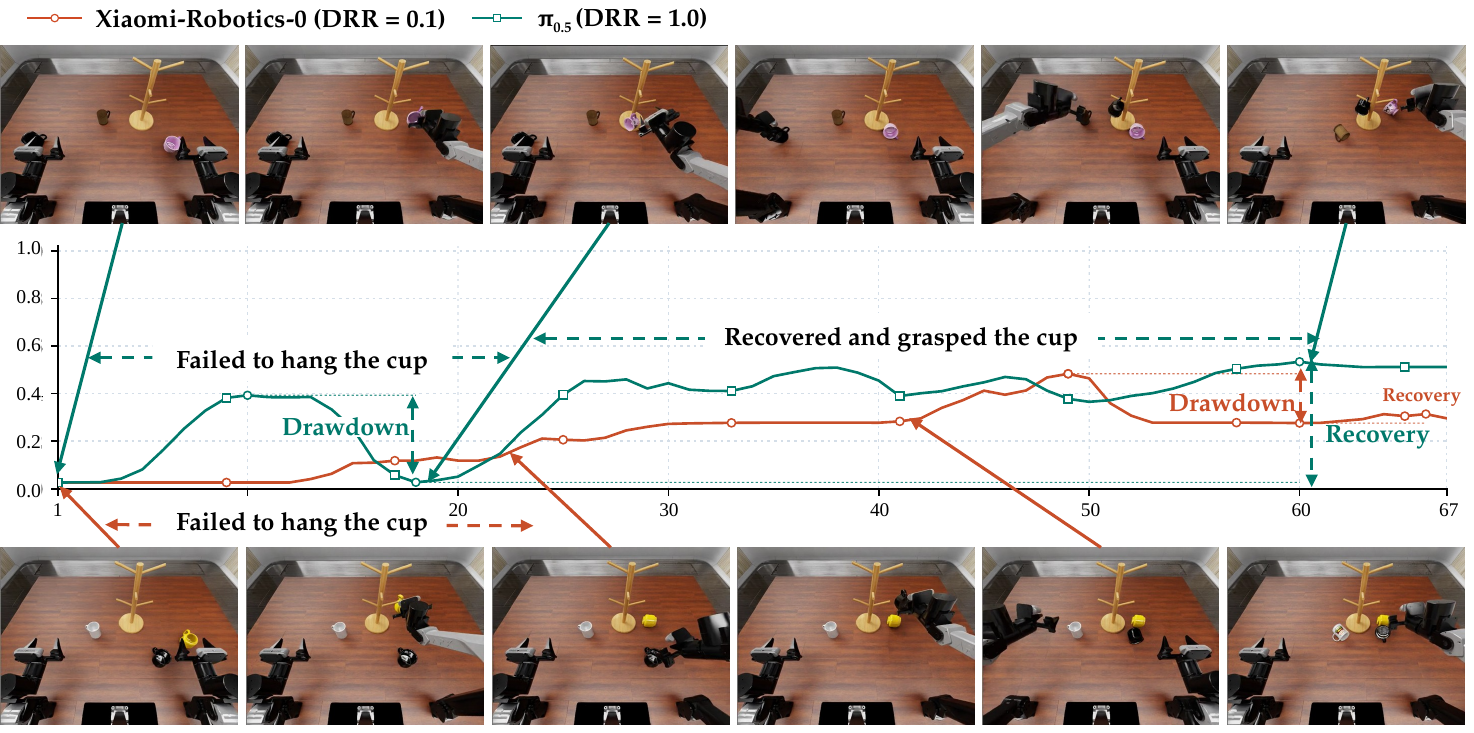}
    \caption{
    \textbf{Drawdown recovery on the \textit{hang mugs} task.}
The upper row shows the $\pi_{0.5}$ rollout, and the lower row shows the Xiaomi-Robotics-0 rollout.
The annotated drawdown and recovery intervals indicate how much progress each policy regains after its maximum-drawdown trough.}
    \label{fig:DRR1}
\end{figure}

\begin{figure}[!b]
    \centering
    \includegraphics[width=\linewidth]{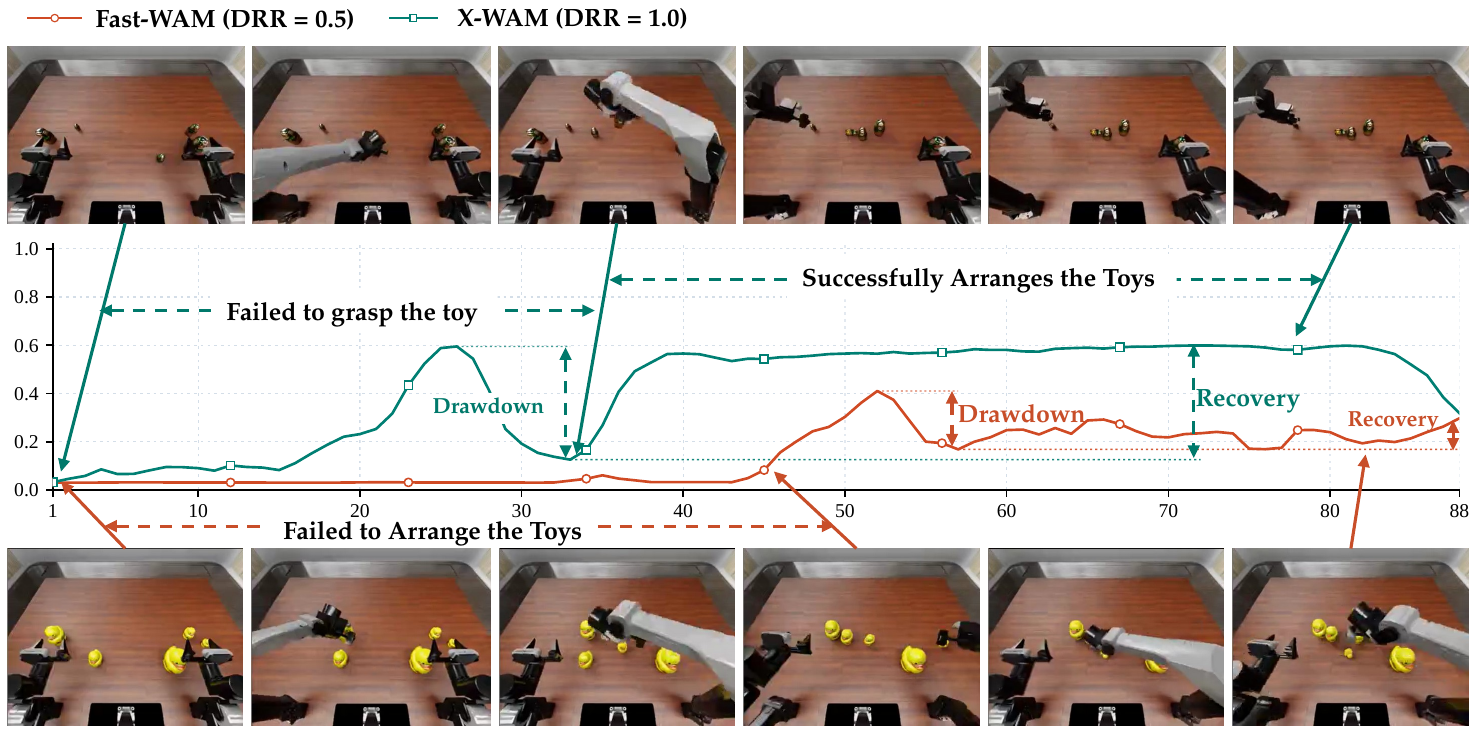}
    \caption{\textbf{Drawdown recovery on the \textit{sort nesting dolls by size} task.}
The upper row shows the X-WAM rollout, and the lower row shows the Fast-WAM rollout.
The annotated drawdown and recovery intervals distinguish sustained post-drawdown recovery from partial recovery.}
    \label{fig:DRR2}
\end{figure}

\paragraph{Drawdown Recovery.}
DRR evaluates how much progress is regained after the trough associated with the maximum drawdown. 
In Figures~\ref{fig:DRR1} and~\ref{fig:DRR2}, the $\pi_{0.5}$-SF rollout first regresses to near-zero progress but subsequently recovers beyond its pre-regression level. 
Xiaomi Robotics 0 also experiences a substantial regression, but fails to regain the lost progress before the rollout terminates, yielding $\mathrm{DRR}\approx0$. 

\clearpage

\begin{figure}[!h]
    \centering
    \includegraphics[width=\linewidth]{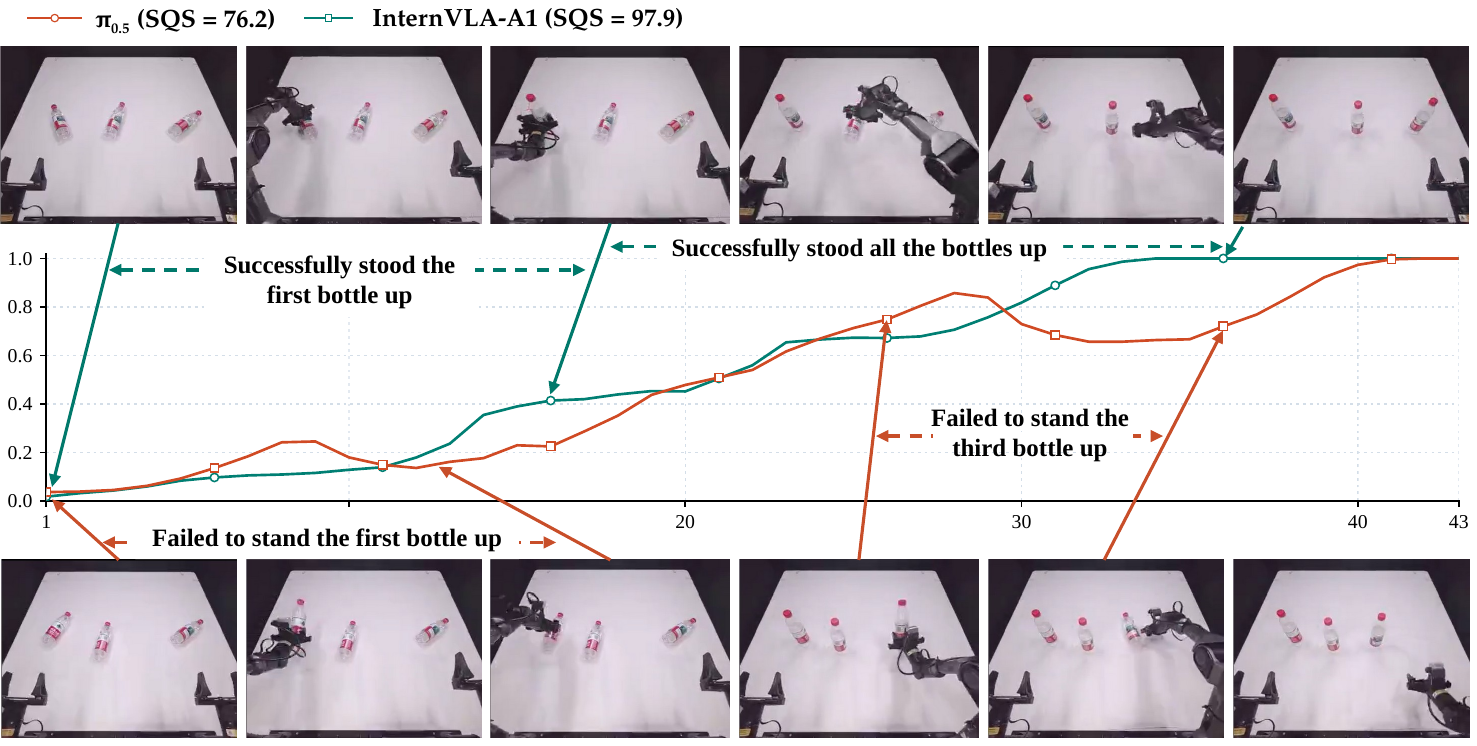}
    \vspace{-20pt}
    \caption{\textbf{Successful-trajectory quality on the \textit{stand up bottles} task.}
The upper row shows the InternVLA-A1 rollout, and the lower row shows the $\pi_{0.5}$ rollout.
The progress curves illustrate how SQS rewards efficient completion while penalizing regression and repeated corrective behavior.}
    \label{fig:SQS1}
\end{figure}

\begin{figure}[!b]
    \centering
    \includegraphics[width=\linewidth]{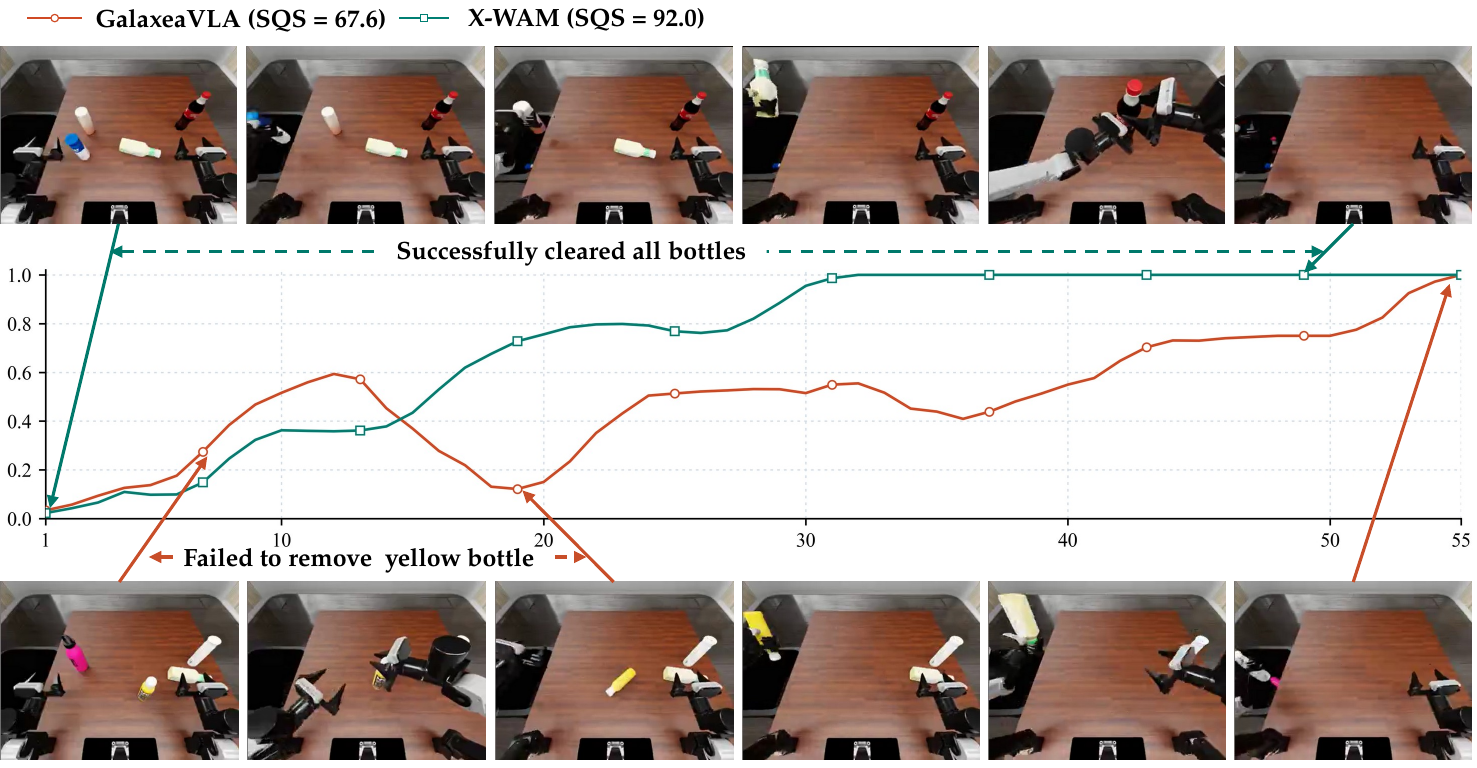}
    \caption{\textbf{Successful-trajectory quality on the \textit{put bottles into dustbin} task.}
The upper row shows the X-WAM rollout, and the lower row shows the GalaxeaVLA rollout.
The progress curves show how SQS distinguishes stable completion from a less efficient trajectory with intermediate regression.}
    \label{fig:SQS2}
\end{figure}

\paragraph{Success Quality.}
SQS evaluates the efficiency and stability of successful executions. 
In Figures~\ref{fig:SQS1} and~\ref{fig:SQS2}, all rollouts eventually succeed, but their execution quality differs. 
InternVLA-A1 and X-WAM make largely sustained progress and complete the task with limited regression.
By contrast, $\pi_{0.5}$ and GalaxeaVLA undergo unsuccessful attempts and a noticeable regression, delaying task completion.

\clearpage

\begin{figure}[h]
    \centering
    \includegraphics[width=\linewidth]{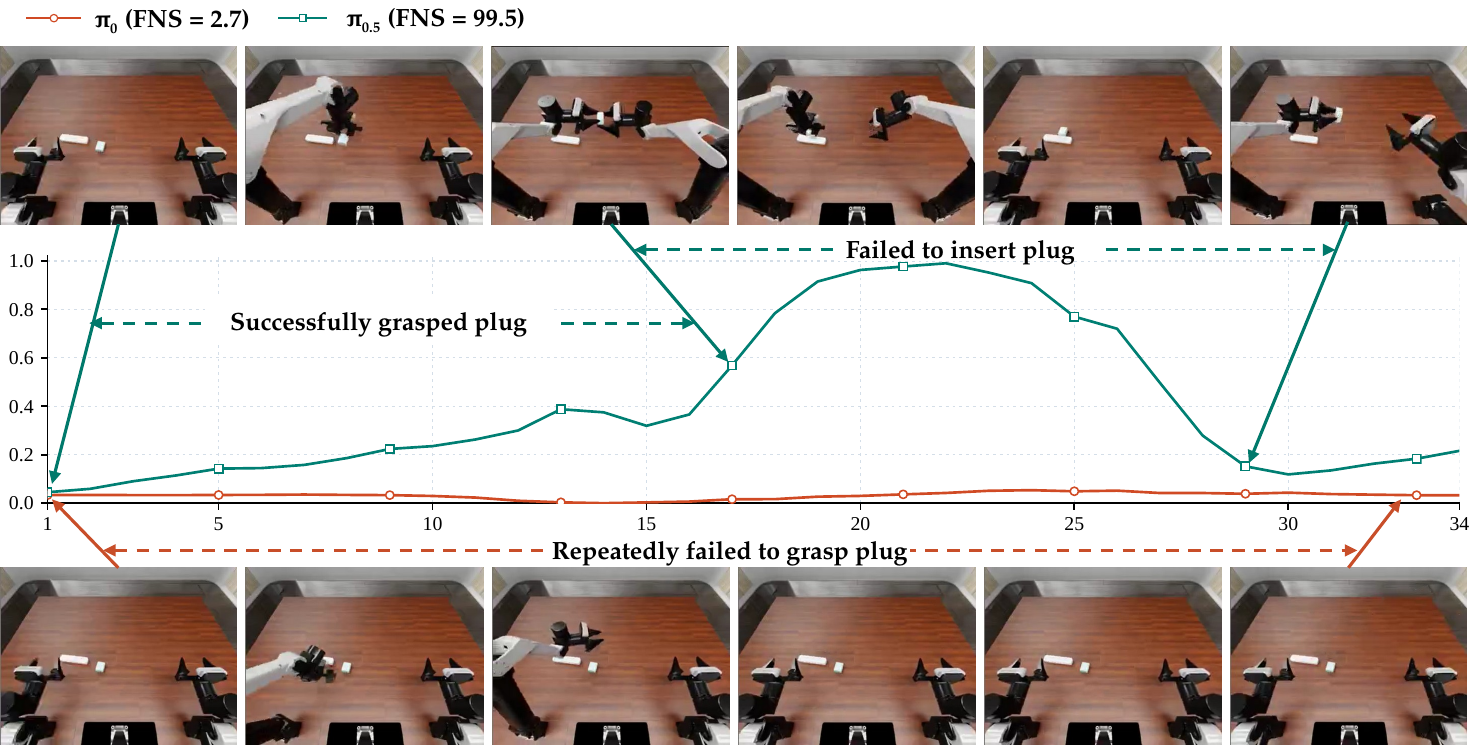}
   \caption{\textbf{Failure proximity on the \textit{plug in charger} task.}
The upper row shows the $\pi_{0.5}$ rollout, and the lower row shows the $\pi_{0}$ rollout.
Although both rollouts fail, FNS assigns a higher score to the trajectory that reaches the late-stage plug-in attempt than to the trajectory that repeatedly fails at grasping.}
    \label{fig:FNS1}
\end{figure}

\begin{figure}[!b]
    \centering
    \includegraphics[width=\linewidth]{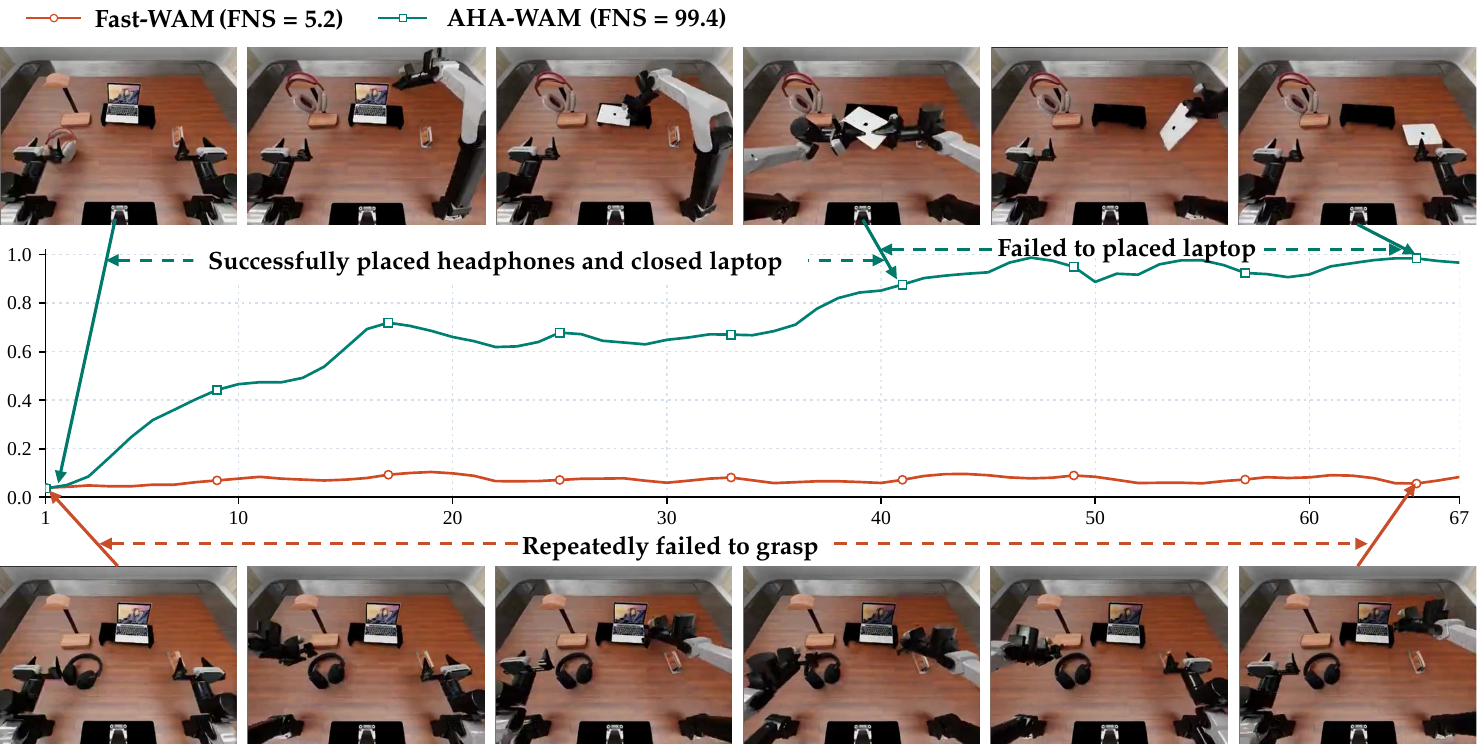}
    \caption{\textbf{Failure proximity on the \textit{store laptop and headphones} task.}
The upper row shows the AHA-WAM rollout, and the lower row shows the Fast-WAM rollout.
FNS differentiates the near-complete trajectory from the rollout that remains stalled at the initial grasping stage.}
    \label{fig:FNS2}
\end{figure}

\paragraph{Failure Near-Success.}
FNS evaluates how close a failed rollout came to completing the task, based on the maximum progress and milestone structure reached before termination.
In Figure~\ref{fig:FNS1} and Figure~\ref{fig:FNS2}, $\pi_{0.5}$ and AHA-WAM successfully complete most of the required manipulation and approach the final goal before failing.
By contrast, $\pi_{0}$ and Fast-WAM repeatedly fail at an early stage.

\clearpage
\subsection{Evaluation Visualization Suite}
\label{subsec:evaluation_visualization}

\begin{figure}[!ht]
\centering
\includegraphics[width=0.97\linewidth]{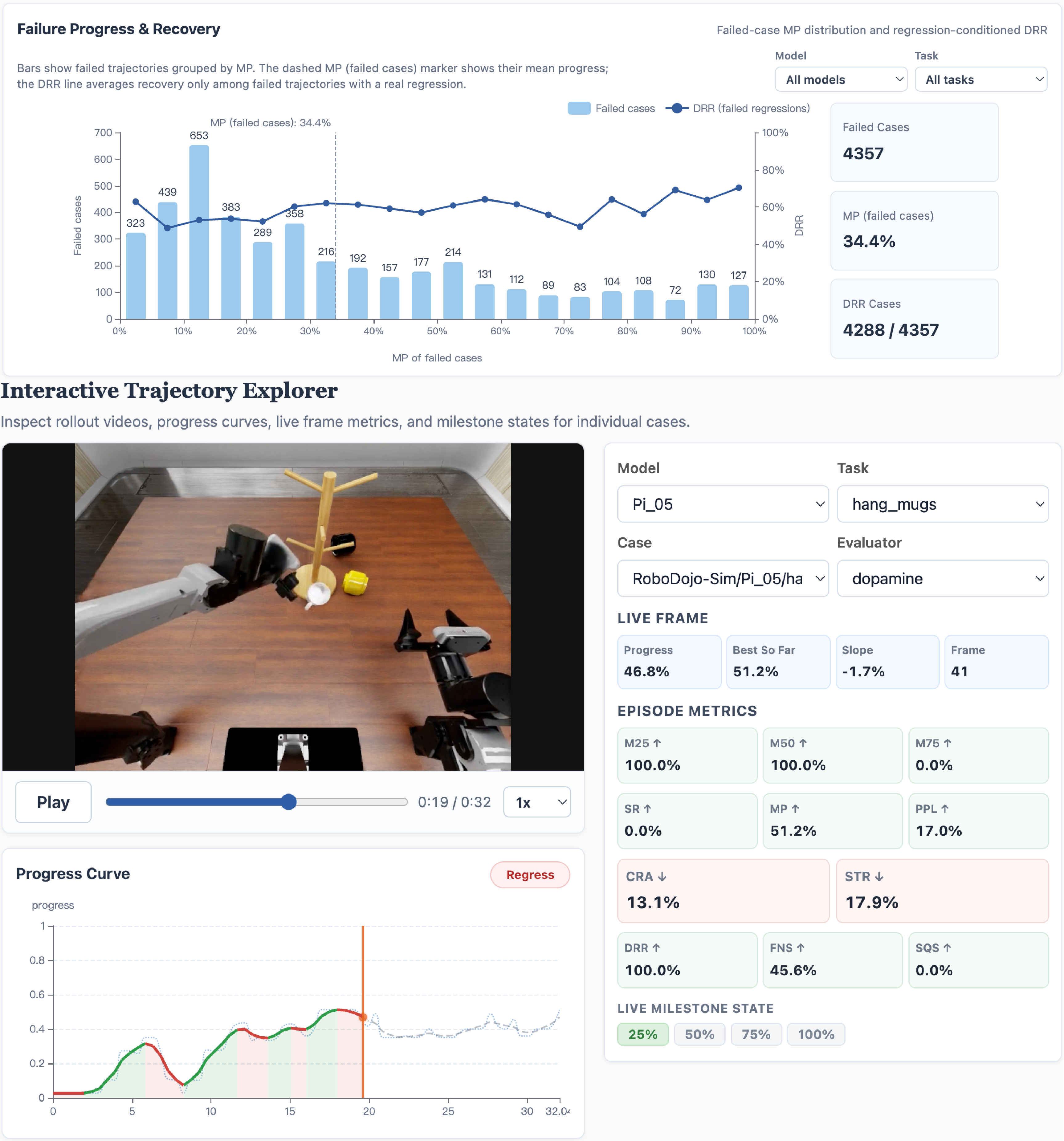}
\caption{\textbf{Automatically Generated Evaluation Report.} PRM-as-a-Judge 1.5 automatically generates a comprehensive report that summarizes model-level results and provides visual analyses of process metrics, success and failure profiles, failure progress, recovery, and individual trajectories.}
\label{fig:evaluation_report}
\end{figure}

PRM-as-a-Judge 1.5 provides a unified visualization suite for presenting and examining trajectory evaluation results, supporting a continuous workflow from result overview to behavioral inspection:
The \textbf{\textit{Model Leaderboard}} offers a compact comparison across evaluated models, while \textbf{\textit{Metric Results}} presents the corresponding evaluation results visually.
\textbf{\textit{Success/Failure Conditional Profiles }}facilitate comparisons between different trajectory outcomes, while \textbf{\textit{Failure Progress and Recovery}} highlight where failures and progress losses occur.
For detailed case inspection, the \textbf{\textit{Interactive Trajectory Explorer}} synchronizes rollout videos with frame-level progress curves, episode-level metrics, and milestone states.
Figure~\ref{fig:evaluation_report} shows a partial view of the web interface.

\end{document}